\documentclass[twoside,twocolumn]{article}

\usepackage{blindtext} 

\usepackage[sc]{mathpazo} 
\usepackage[T1]{fontenc} 
\usepackage{microtype} 

\usepackage[english]{babel} 

\usepackage[margin=0.5in,top=32mm,columnsep=20pt]{geometry} 
\usepackage[hang, small,labelfont=bf,up,textfont=it,up]{caption} 
\usepackage{booktabs} 

\usepackage{lettrine} 

\usepackage{enumitem} 
\setlist[itemize]{noitemsep} 

\usepackage{abstract} 

\usepackage{titlesec} 
\renewcommand\thesection{\Roman{section}} 
\renewcommand\thesubsection{\roman{subsection}} 
\titleformat{\section}[block]{\large\scshape\centering}{\thesection.}{1em}{} 
\titleformat{\subsection}[block]{\large}{\thesubsection.}{1em}{} 

\usepackage{fancyhdr} 
\usepackage{titling} 

\usepackage{hyperref} 
\usepackage{amsopn}
\usepackage{url}
\usepackage{amsmath}
\usepackage{dsfont}
\usepackage{float}
\usepackage{color}
\usepackage{algorithmic}

\usepackage{graphicx,epstopdf}
\usepackage{braket,amsfonts}
\usepackage[caption=false]{subfig}
\DeclareMathOperator{\Radon}{\mathcal{R}}
\DeclareMathOperator{\Genop}{\mathcal{A}}

\pretitle{\begin{center}\Huge\bfseries} 
\posttitle{\end{center}} 
\title{On the crucial impact of the coupling projector-backprojector in iterative tomographic reconstruction} 
\author{%
\textsc{Filippo Arcadu$^{*,\dagger}$, Marco Stampanoni$^{*,\dagger}$ and Federica Marone$^{\dagger}$}
\\[1ex] 
\normalsize $^{(*)}$Institute for Biomedical Engineering, ETH Zurich, 8092 Zurich, Switzerland \\
\normalsize $^{(\dagger)}$Swiss Light Source, Paul Scherrer Institute, 5232 Villigen, Switzerland \\ 
\normalsize \href{mailto:filippo.arcadu@gmail.com}{filippo.arcadu@gmail.com}\\ 
\normalsize \href{mailto:https://github.com/arcaduf}{https://github.com/arcaduf}\\
\normalsize 16th December 2016 
}
\normalsize\date{} 

\renewcommand{\maketitlehookd}{%
\begin{abstract}
\noindent
The performance of an iterative reconstruction
algorithm for X-ray tomography is strongly determined by the features of the used
forward and backprojector.
For this reason, a large number of studies has focused on the to design of projectors with increasingly higher accuracy and speed. 
To what extent the accuracy of an iterative algorithm is affected by the
mathematical affinity and the similarity between the actual implementation of the forward
and backprojection, referred here as ``coupling projector-backprojector'', has been an overlooked aspect so far.
The experimental study presented here shows that the reconstruction quality and the convergence
of an iterative algorithm greatly rely on a good matching between the implementation of the tomographic operators.
In comparison, other aspects like the accuracy of the standalone operators, the usage of physical constraints or 
the choice of stopping criteria may even play a less relevant role.
\end{abstract}
\centering
{\fontsize{9}{2}\selectfont\textbf{Keywords:} X-ray tomography, iterative tomographic reconstruction, forward projector, backprojector.}
}

\begin{document}

\maketitle


\section{Introduction}
\label{sec:intro}
\lettrine[nindent=0em,lines=3]{I}terative reconstruction for X-ray tomography has been studied since the introduction of the first CT scans in the mid 70s \cite{Hounsfield1973}.
Differently from the filtered backprojection (FBP) algorithm \cite{Herman2009}, iterative methods are non-linear and less computationally efficient,
as the forward projector and its adjoint operator, the backprojector, are generally called few times per iteration.
In contrast to FBP, iterative methods can, however, provide high quality reconstructions of tomographic underconstrained datasets, 
characterized by poor signal-to-noise ratio (SNR), little number of views and/or missing data.

In general, iterative algorithms consist of the following elements: a solver for the cost function, physical constraints,
a regularization scheme linked to the a-priori-knowledge regarding the specimen under study and tomographic projectors.

Four main families of solvers can be identified for iterative reconstruction. Algebraic reconstruction techniques like ART \cite{Herman1973}, SIRT
\cite{Gilbert1972} and 
SART \cite{Andersen1984} handle the tomographic problem as a system of equations, which is solved by means of the Kaczmarz method \cite{Kaczmarz1937}. 
Statistical methods as the maximum likelihood expectation maximization (MLEM) \cite{Shepp1982}, the separable paraboloidal surrogate \cite{Erdogan1999} and 
the penalized weighted least square method (PWLS) \cite{Fessler1994,Elbakri2002} incorporate the statistical model ruling the signal formation at the detector.
Recently, modern techniques for convex optimization like the split Bregman method \cite{Goldstein2009}
and the alternate direction method of multipliers (ADMM) \cite{Boyd2010} have also been applied to tomographic reconstruction
\cite{Wang2012,Ramani2012,Chun2014}. Finally, the projection-onto-convex-sets method \cite{Defrise2006} has been mainly used to address the interior tomography
problem.

Physical constraints enforce at each iteration strict conditions in the image domain. Setting to zero all negative pixel values and those falling outside the 
reconstruction circle  is a typical example of broadly exploited physical constraints.

Regularization schemes often utilized by iterative algorithms are Tikhonov \cite{Tikhonov1977}, Huber \cite{Huber1964} and total variation (TV) \cite{Rudin1992}.
In particular, a Huber or TV term can steer the cost function towards a piecewise-constant solution, while preserving the spatial resolution.

Several implementations of the tomographic projectors have been proposed since the 70s. The pixel-driven \cite{Herman2009,Peters1981,Zhuang1994},
ray-driven \cite{Herman2009,Zhuang1994,Zeng1993}, distance-driven \cite{Basu2002,Basu2004} and slant-stacking \cite{Chapman1981,Toft1996} approaches are
different methods to approximate the Radon transform in real domain. Since the listed approaches feature a complexity of $\mathcal{O}(N^{3})$ \cite{Toft1996}, 
their implementation on GPUs is a must to make iterative reconstructions computationally feasible \cite{YongLong2010,Palenstijn2011,Papenhausen2013,Andersson2016}. 
Tomographic projectors with complexity $\mathcal{O}(N^{2}\log_{2}N)$, based on hierarchical-decomposition \cite{Basu2000},
the non-uniform Fourier transform \cite{Matej2004} or the gridding method \cite{Arcadu2016}, are, instead, fast enough to not necessarily require 
a GPU architecture.

So far, research in iterative reconstruction algorithms has mainly addressed the design of regularization schemes leading to a
better SNR-spatial resolution tradeoff and 
the development of tomographic projectors with increasingly higher accuracy and speed.
An aspect that has been generally neglected is the role of the \textit{coupling projector-backprojector}, i.e., the level of mathematical affinity and 
matching between the actual implementation of the forward projector and its adjoint operator.

This work is an empirical investigation of the role played by this aspect on the performance of iterative reconstruction algorithms for X-ray tomography.
Ad-hoc experiments with state-of-the-art implementations of different tomographic operators
(pixel-driven, ray-driven, distance-driven, slant-stacking, gridding method) have been designed for this purpose.
Reconstructions have been performed with both analytical (FBP) and iterative (ADMM, PWLS, MLEM, SIRT) schemes.
Results show that the coupling projector-backprojector substantially affects accuracy and convergence of an iterative algorithm.
In some cases, the degree of matching between the tomographic projectors can even play a more decisive role for the performance of the iterative
method than other factors, like physical constraints, stopping criteria or the accuracy of the standalone projectors.

A mathematical justification of the presented experimental results is not straightforward and is outside the scope of this work.
The aim of this study is, instead, to provide convincing experimental evidence that a well-tuned coupling 
projector-backprojector is an absolute ``must'' for iterative tomographic reconstruction schemes to avoid systematically sub-accurate results. 
A practical strategy for measuring the coupling degree is also proposed: this tool could be very useful for users and developers of software packages
for iterative tomographic reconstruction to assess and validate the quality of the proposed projector pairs.

\section{Experimental framework}

\subsection{Tomographic projectors}
\label{tomographic-projectors}
 The Radon transform, $\Radon$, integrates a function $f(\mathbf{x})\::\,\mathbb{R}^{n} \longrightarrow \mathbb{R}$ 
over an hyperplane $HY(\mathbf{n},t) = \{\mathbf{x} \in \mathbb{R}^{n}\:|\: \mathbf{x}\cdot\mathbf{n} = t\}$, where
$\mathbf{n}$ is a unit vector and $t \in \mathbb{R}$ is the signed distance from the origin \cite{Natterer2001}:
\begin{equation}
\begin{split}
  \Radon\{f\}(\mathbf{n},t) &:= \int\limits_{HY}d\mathbf{x}\:f(\mathbf{x})
                                      = \int\limits_{\mathbb{R}^{n}}d\mathbf{x}\:\delta(t-\mathbf{x}\cdot\mathbf{n})f(\mathbf{x})\\
                                      &= \int\limits_{\mathbf{n}^{\bot}}d\mathbf{x}\:f(t\mathbf{n} + \mathbf{x}) \hspace{0.3cm}.
  \label{radon_transform_1}
\end{split}
\end{equation}
$\delta$ is the Dirac function and 
$\mathbf{n}^{\bot} = \{\mathbf{x} \in \mathbb{R}^{n}\:|\:\mathbf{x}\cdot\mathbf{n} = 0\}$ is the subspace
orthogonal to $\mathbf{n}$.
For $n=2$, $f(\mathbf{x}) = f(x_{1},x_{2})$, $\mathbf{n} = (\cos\theta,\sin\theta)$, 
$HY$ is a line of equation $\mathbf{x}\cdot\mathbf{n} = x_{1}\cos\theta + x_{2}\sin\theta = t$,
thus, $\Radon$ integrates $f$ along lines.
The second definition in (\ref{radon_transform_1}) simplifies to:
\begin{equation}
    \Radon\{f\}(\theta,t) := \int\limits_{-\infty}^{+\infty}dx_{1}\int\limits_{-\infty}^{+\infty}dx_{2}\:f(x_{1},x_{2})\:
                             \delta(x_{1}\cos\theta + x_{2}\sin\theta - t) \hspace{0.3cm}. 
    \label{radon_transform_2}
\end{equation}
$\Radon$ is also called forward projector and $\theta$ is, here, the angle formed by the detector line and the positive $x$-semiaxis.
The dual transform, i.e., the adjoint of the Radon transform, $\Radon^{*}$, is called
backprojection. For $n=2$ and given a generic function $g(\mathbf{x}) = g(x_{1},x_{2})$, $\Radon^{*}$ is defined as \cite{Natterer2001}:
\begin{equation}
  \Radon^{*}\{ g\}(\mathbf{x}) = \frac{1}{2\pi} \int\limits_{0}^{2\pi} d\theta \: g\left( \theta , x_{1}\cos\theta + x_{2}\sin\theta \right) \hspace{0.5cm}.
  \label{radon_transform_3}
\end{equation}
The six implementations of $\Radon$ and $\Radon^{*}$ used in this work are for parallel beam geometry and a brief description is given in the following.
\newline
The pixel-driven (PD) approach \cite{Herman2009,Peters1981,Zhuang1994}
\begin{figure*}[!t]
  \centering
    \hspace*{-0.4cm}
    \subfloat[Pixel-driven approach]{{\includegraphics[width=5cm]{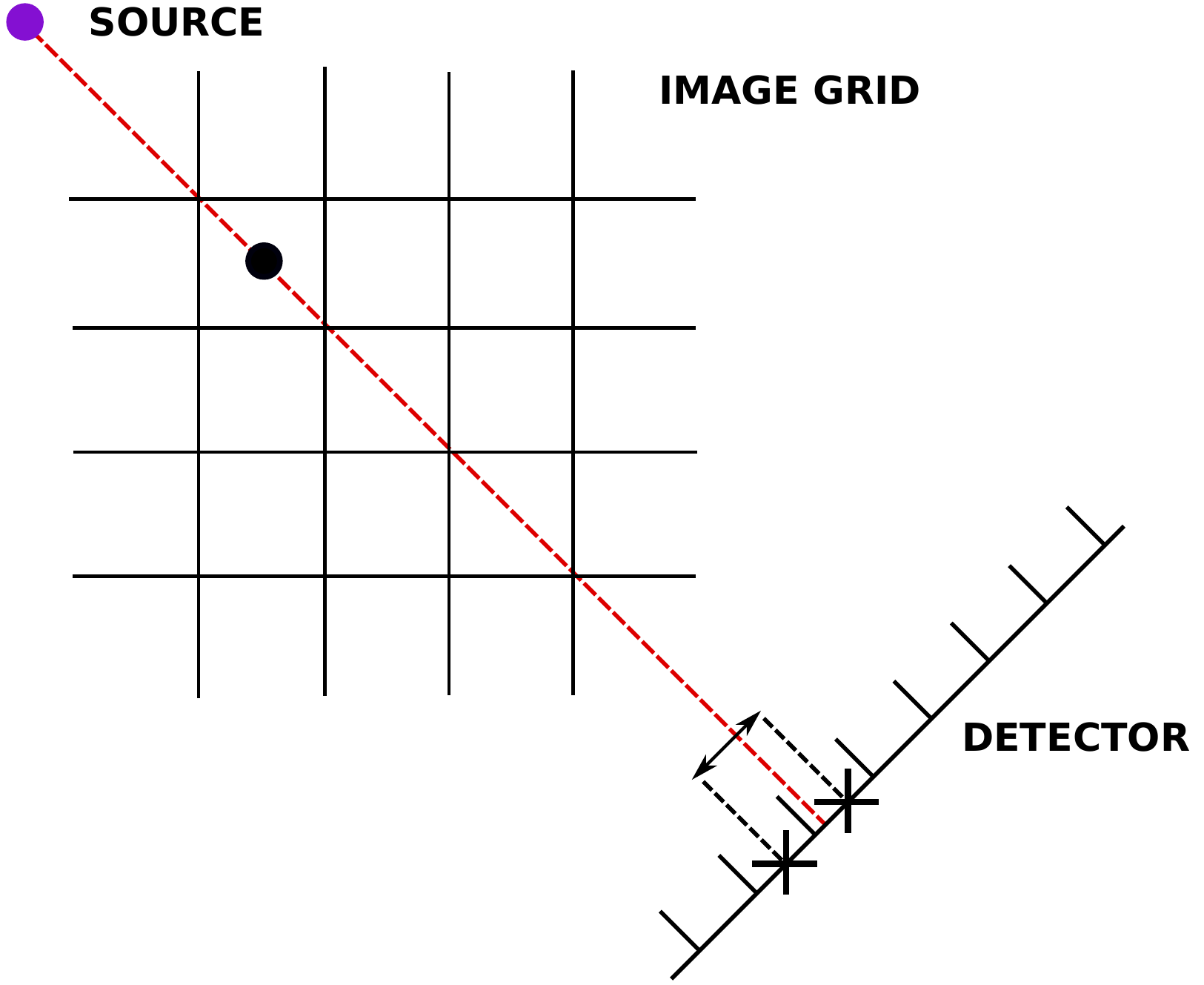}\label{radon-transform-pixel} }}%
    \hspace*{0.0cm}\subfloat[Ray-driven approach]{{\includegraphics[width=5cm]{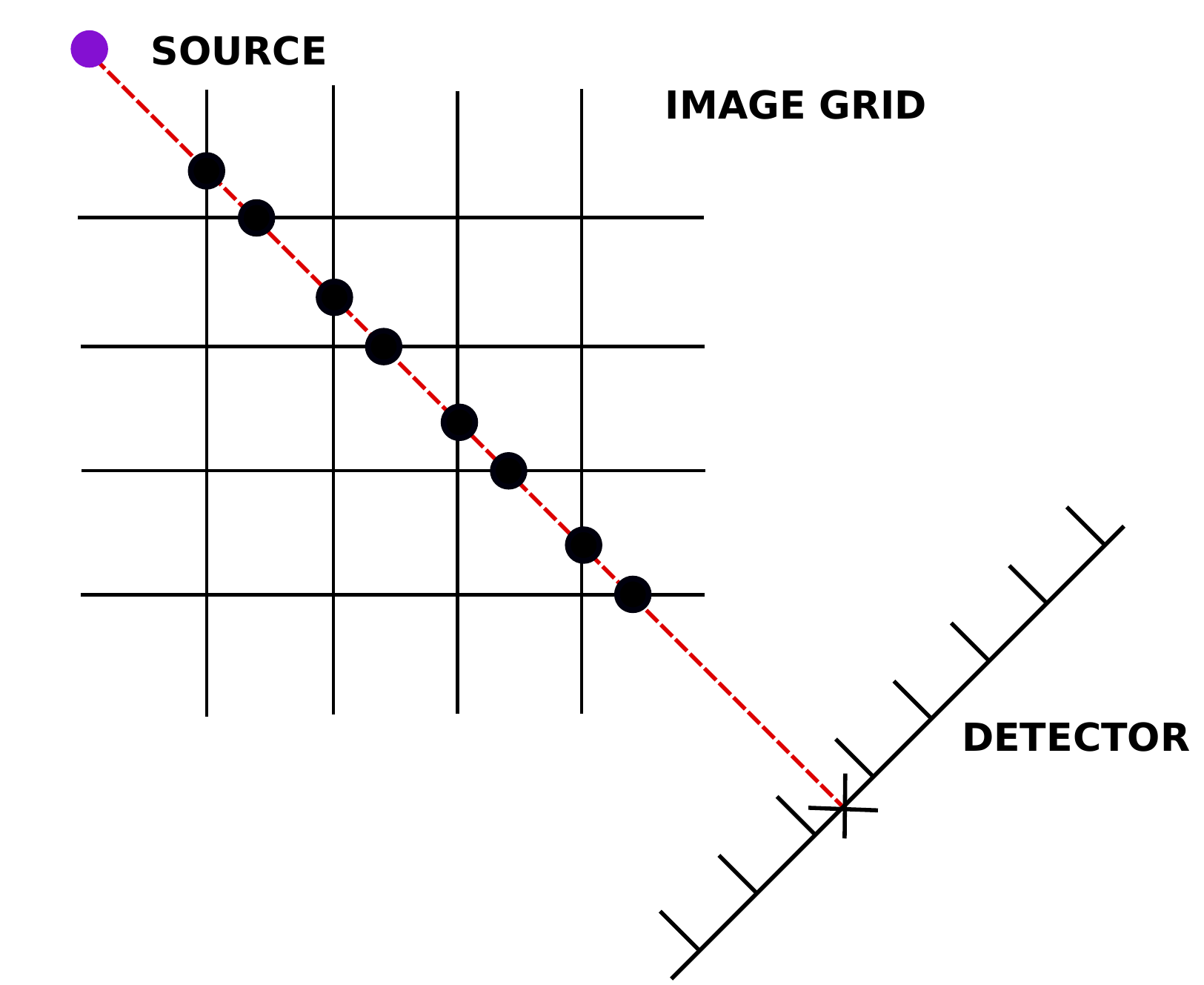}\label{radon-transform-ray}  }}%
    \subfloat[Distance-driven approach]{{\includegraphics[width=5cm]{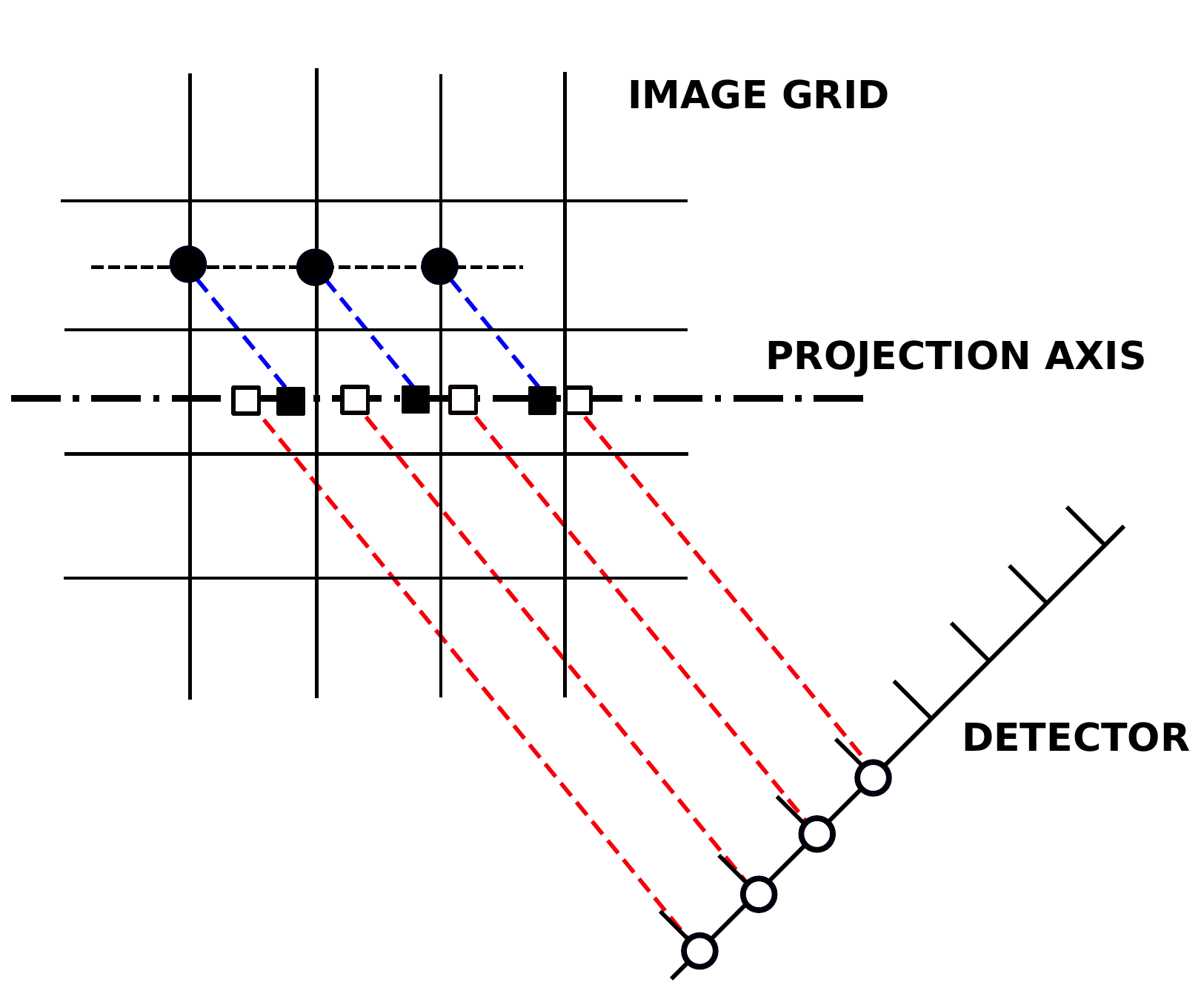}\label{radon-transform-distance}  }}%
    \hspace*{0.0cm}\subfloat[Slant-stacking approach]{{\includegraphics[width=5cm]{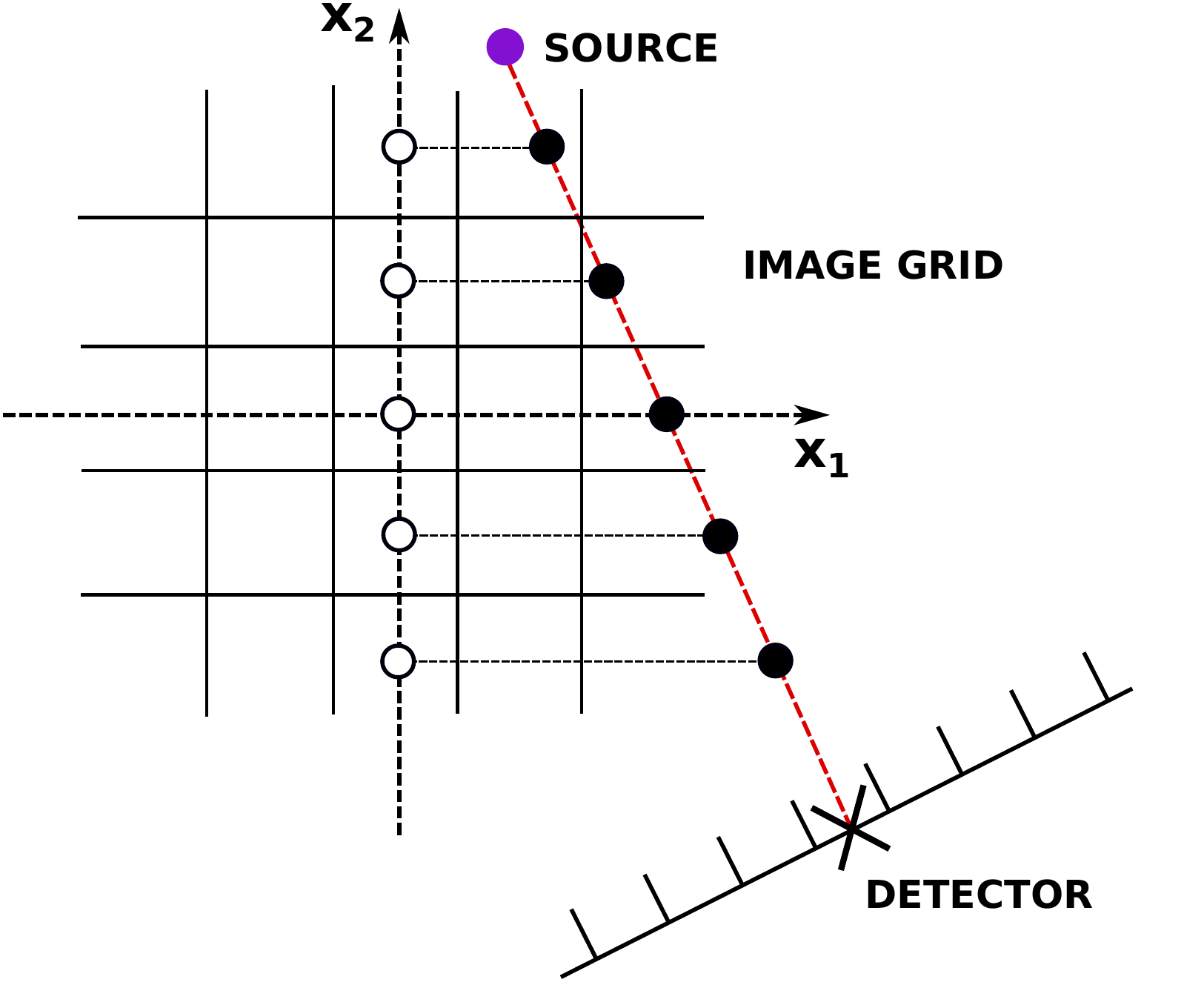}\label{radon-transform-slant}  }}%
    \caption{Schematic representation of the different mechanisms characterizing the pixel-driven,
              ray-driven, distance-driven and slant-stacking approach for forward projection (and backprojection).}%
    \label{radon-transform-aaproaches}%
\end{figure*}
works by connecting the source point to the selected pixel center until intersection
with the detector line, as displayed in Fig.\ref{radon-transform-pixel}. The pixel value is assigned on the basis of a linear
interpolation scheme to the two detector cells that enclose the ray end point (they are indicated with a cross in Fig.\ref{radon-transform-pixel}).

The ray-driven (RD) approach \cite{Herman2009,Zhuang1994,Zeng1993} connects the source to the center of a selected detector cell 
as shown in Fig.\ref{radon-transform-ray}. The Siddon algorithm \cite{Siddon1985} is used to compute the intersection points of the ray 
with the image grid (black dots in Fig.\ref{radon-transform-ray}). Each pixel contributes to the selected detector cell according to the ray path length.

The distance-driven (DD) approach \cite{Basu2002,Basu2004} in Fig.\ref{radon-transform-distance}
projects the pixel boundaries (black dots) of each image row/column and
the detector cell boundaries (white dots) onto a common axis (in Fig.\ref{radon-transform-distance}, 
the black squares are projected pixel boundaries, the white squares projected detector cells).
The overlap between the interval defined by the projected boundaries of an image pixel and the one defined by the projected boundaries
of a detector cell weights the contribution of the selected image pixel to the selected detector cell (and viceversa).

The slant stacking (SS) \cite{Toft1996} method connects the source to each detector cell and divides the interval $[0,\pi]$
in two regions: one for nearly-vertical lines $0 \leq \theta \leq \pi/4$ and $3\pi/4 \leq \theta \leq \pi$;
one for nearly-horizontal lines $\pi/4 \leq \theta \leq 3\pi/4$ ($\theta$ is the angle formed by the detector line and 
the positive $x$-semiaxis). Figure \ref{radon-transform-slant} shows how SS works
with a nearly-vertical line: the abscissas of the black dots are obtained by using the ray equation
and the $x_{2}$ coordinates (white dots) of all image pixels. The computed points $(x_{1},x_{2})$
contribute to the selected detector cell according to a linear interpolation scheme. The same approach is used for nearly-horizontal lines.

The gridding projectors \cite{Matej2004,Arcadu2016} are implementations of $\Radon$ and $\Radon^{*}$ in the Fourier domain and 
are based on the Fourier slice theorem (FSM) \cite{Natterer2001}. For the forward operation, the input image grid is, first, 
multiplied with the deapodization matrix and, then, Fourier transformed (FFT-2D). The Fourier Cartesian grid is convolved with a compact kernel to
obtain Fourier samples on a polar grid. According to the FSM, the inverse Fourier transform (IFFT-1D) of a polar slice at angle $\theta$ corresponds
to the object projection acquired at angle $\theta$. The accuracy and efficiency of gridding projectors rely entirely on the choice of 
the convolving kernel (that also determines the deapodizer) and the oversampling ratio, $\alpha$, used for the Fourier grid.
In this work, two slightly different implementations are considered \cite{Arcadu2016}: one using a prolate spheroidal wavefunctions kernel and $\alpha=2$ (abbreviated with WF);
the other using a Kaiser-Bessel kernel and $\alpha = 1.5$ (abbreviated with KB).

\subsection{Degree of coupling projector-backprojector}
Given a generic linear operator $\Genop\::\:\mathbb{C}^{n_{1}}\longrightarrow\mathbb{C}^{n_{2}}$, 
the adjoint, $\Genop^{*}$, is defined as follows:
\begin{equation}
\begin{split}
  &\Genop^{*}\::\:\mathbb{C}^{n_{2}}\longrightarrow\mathbb{C}^{n_{1}} \hspace{0.5cm}\text{such that}\hspace{0.5cm}
  \left< \mathbf{y} , \Genop(\mathbf{x}) \right> = \left< \Genop^{*}(\mathbf{y}) , \mathbf{x} \right>\\ 
  &\hspace{3.0cm} \forall\:\mathbf{x} \in \mathbb{C}^{n_{1}} \,,\, \forall\:\mathbf{y} \in \mathbb{C}^{n_{2}}\hspace{0.5cm},
  \label{adjoint-definition}
\end{split}
\end{equation}
where $<...>$ is the notation for the inner product. Definition (\ref{adjoint-definition}) can be used to measure how well a
computer implementation of $\Genop$ matches the computer implementation of $\Genop^{*}$. 
The two inner products in (\ref{adjoint-definition}) are numerically
evaluated with $\mathbf{x}$ and $\mathbf{y}$ being vectors of randomly generated numbers.
If the ratio $r = \left< \Genop^{*}(\mathbf{y}) , \mathbf{x} \right>/\left< \mathbf{y} , \Genop(\mathbf{x}) \right>$
matches 1 up to a reasonably sufficient numerical precision, 
the implementations of $\Genop$ and $\Genop^{*}$ can be considered well coupled.

For the tomographic case, a good coupling is achieved when the backprojector foresees the same exact operations of the forward projector,
but in reverse order and switching the roles of input/output arrays for object and sinogram. The coupled implementations of 
$\Radon$ and $\Radon^{*}$ listed in (\ref{tomographic-projectors}) feature $r = 1$ up to the 7th digit. 
When not coupled, $r = 1$ at most up to the 4th digit. 

\subsection{Reconstruction algorithms}
Analytical reconstructions are here performed with filtered backprojection (FBP) \cite{Herman2009}, that inverts the Radon transform by applying the 
linear operator $\Radon^{*} \circ\: \Delta$, where $\Delta$ is the ramp or Ram-Lak filter.
The tradeoff between SNR and spatial resolution of FBP reconstructions depends on the type of window superimposed to the Ram-Lak filter \cite{Kak2001}.
For this reason, FBP is used here with four different filters \cite{Lyra2011}: a pure Ram-Lak filter that provides the highest spatial resolution and poorest SNR 
(abbr. RAMP); a Ram-Lak filter combined with a Shepp-Logan window (abbr. SHLO); a Ram-Lak filter combined with a Hanning window 
(abbr. HANN); a Ram-Lak filter combined with a Parzen window that provides the poorest spatial resolution and highest SNR (abbr. PARZ).

Four different iterative reconstruction algorithms have been selected for this study:
the alternate direction method of multipliers (ADMM) with TV
regularization \cite{Ramani2012}, the penalized weighted least square (PWLS) with Huber penalty \cite{Fessler1994,Elbakri2002},
the maximum-likelihood expectation maximization (MLEM) \cite{Shepp1982} and the simultaneous 
iterative reconstruction technique (SIRT) \cite{Herman1973}.
The number of iterations is set to around 100, when studying the algorithm convergence. For the other experiments,
the stopping criterion and regularization strength are optimized according to the characteristics of the considered dataset.
Iterative reconstructions are run with a range of different stopping criteria and weights of the penalty term.
We define the optimal number of iterations and regularization strength as those providing the best reconstruction accuracy,
after appropriate exploration of the parameter space. Nevertheless, it is important to point out that 
the presented trends in the performance of the iterative algorithms as a function of the coupling projector-backprojector are
independent from the choice of the regularization parameters and confirmed also in case of a suboptimal selection.

\subsection{Dataset and image quality assessment}
The Shepp-Logan (SL) phantom \cite{Shepp1974} is used to create the simulated datasets for this study.
Since this phantom consists exclusively of roto-translated ellipses, its forward projection can be computed analytically \cite{Kak2001}.
An analytical forward projection can be used in two ways: (i) as reference when measuring the accuracy of a projector;
(ii) as tomographic dataset not coupled to a specific operator used within the selected reconstruction algorithm.
A selection of experiments presented in Section \ref{operator-coupling-analytical-reconstruction} and \ref{operator-coupling-iterative-reconstruction}
were also performed with different simulated objects and real datasets:
the observed trends are comparable to those obtained with the SL phantom and are, therefore, independent from the chosen object.

The discretized forward projection of an object is also called sinogram, which corresponds to a matrix $\, \in \mathbb{R}^{M\times N}$; $M$
is the number of views and $N$ the number of detector cells. In this study, projections are always
homogeneously distributed in $[0,\pi)$.
A sinogram in parallel beam geometry is considered undersampled,
when $M < N \pi/2$ \cite{Kak2001}. FBP reconstructions of undersampled datasets are affected by radially arranged line artifacts \cite{Kak2001}.
To simulate projections with a low photon statistics, Poisson noise with variance $\sigma$ is added to the computed forward projection.
Poissonian statistics accounts for the shot noise affecting real projection data, whereas it neglects other sources
of signal corruption, e.g., roundoff errors and electrical noise, not considered here.

Four different analytical forward projections of the SL phantom are used in the experimental sections:
a well-sampled, noiseless SL sinogram with 402 views $\times$ 256 pixels, abbreviated as SL-FULL;
an undersampled, noiseless SL sinogram with 50 views $\times$ 256 pixels, abbreviated as SL-UNDER;
a well-sampled, noisy SL sinogram with 402 views $\times$ 256 pixels and additional Poisson noise with $\sigma=3\%$ of the SL-FULL mean value,
abbreviated as SL-NOISE; an undersampled noisy sinogram with 75 views $\times$ 256 pixels and additional Poisson noise with the same $\sigma$
of the SL-NOISE, abbreviated as SL-UCONSTR.

The image quality is measured by the peak-signal-to-noise ratio (PSNR) \cite{HuynhThu2008}, defined as:
\begin{equation}
  \text{PSNR} = 10\:\log_{10}\left( \frac{\max\{\mathbf{r}\}^{2}}{\text{MSE}} \right) =
                20\:\log_{10}\left( \frac{|\max\{\mathbf{r}\}|}{\sqrt{\text{MSE}}} \right) \hspace{0.5cm},
  \label{psnr-formula} 
\end{equation}
where the mean squared error (MSE) is:
\begin{equation}
  \text{MSE} = \frac{1}{PQ}\sum\limits_{i = 0}^{P-1}\sum\limits_{j = 0}^{Q-1}\left( f[i,j] - r[i,j] \right)^{2} \hspace{0.5cm}. 
  \label{mse-formula}
\end{equation}
$\mathbf{r}, \mathbf{f} \in \mathbb{R}^{P\times Q}$ are the reference and the image to be evaluated, respectively. 
The PSNR is preferable over the MSE because more sensitive: as $( f[i,j] - r[i,j] )^{2}$ appears at the 
denominator, even small differences can elicit non negligible variations of the PSNR value.
In this study, the reference is either SL or its analytical forward projection.
When comparing an analytical or iterative reconstruction to SL, the PSNR is computed within the
reconstruction circle.

\section{Operator coupling in analytical reconstruction}
\label{operator-coupling-analytical-reconstruction}
The following FBP tests provide a first indication of the role played by the coupling
projector-backprojector in iterative tomographic reconstruction. Reconstructed slices are not displayed here,
because differences are usually not detectable at visual inspection.

\begin{figure*}[!t]
  \centering
   \subfloat[SL-FULL]{{\includegraphics[width=6cm]{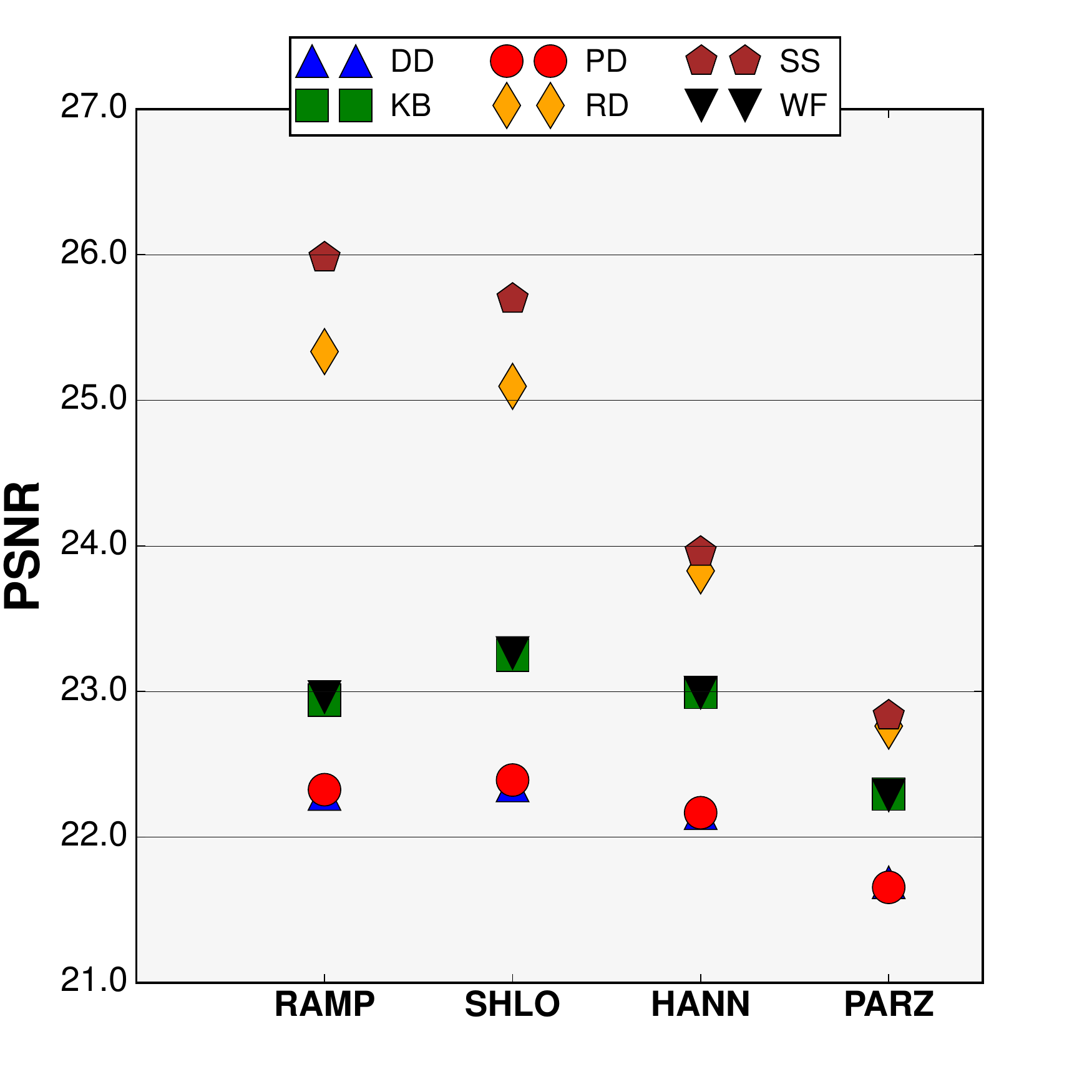}\label{anal-backproj-perfect} }}%
   \hspace*{0.5cm}\subfloat[SL-UNDER]{{\includegraphics[width=6cm]{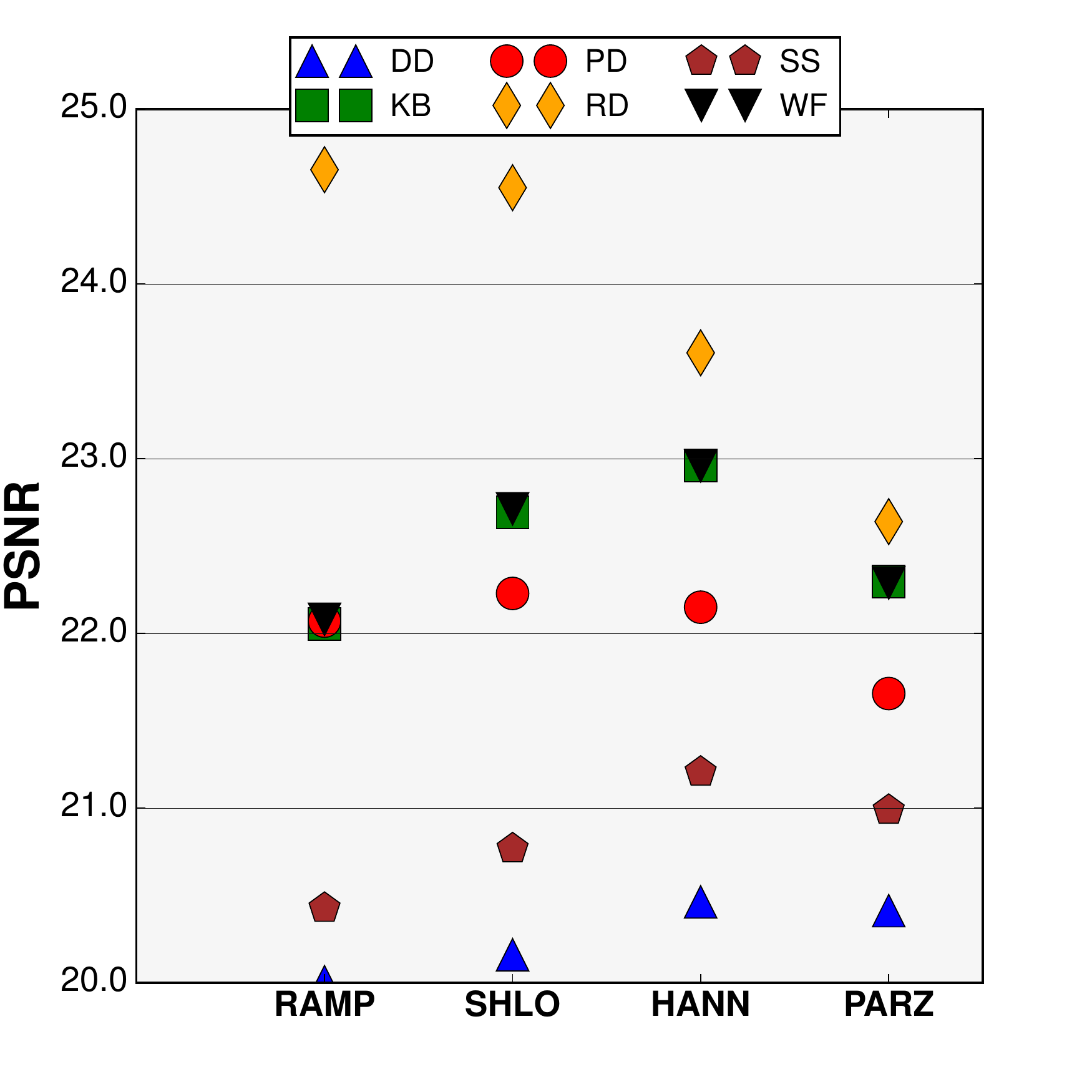}\label{anal-backproj-under}  }}%
   \hspace*{0.5cm}\subfloat[SL-UCONSTR]{{\includegraphics[width=6cm]{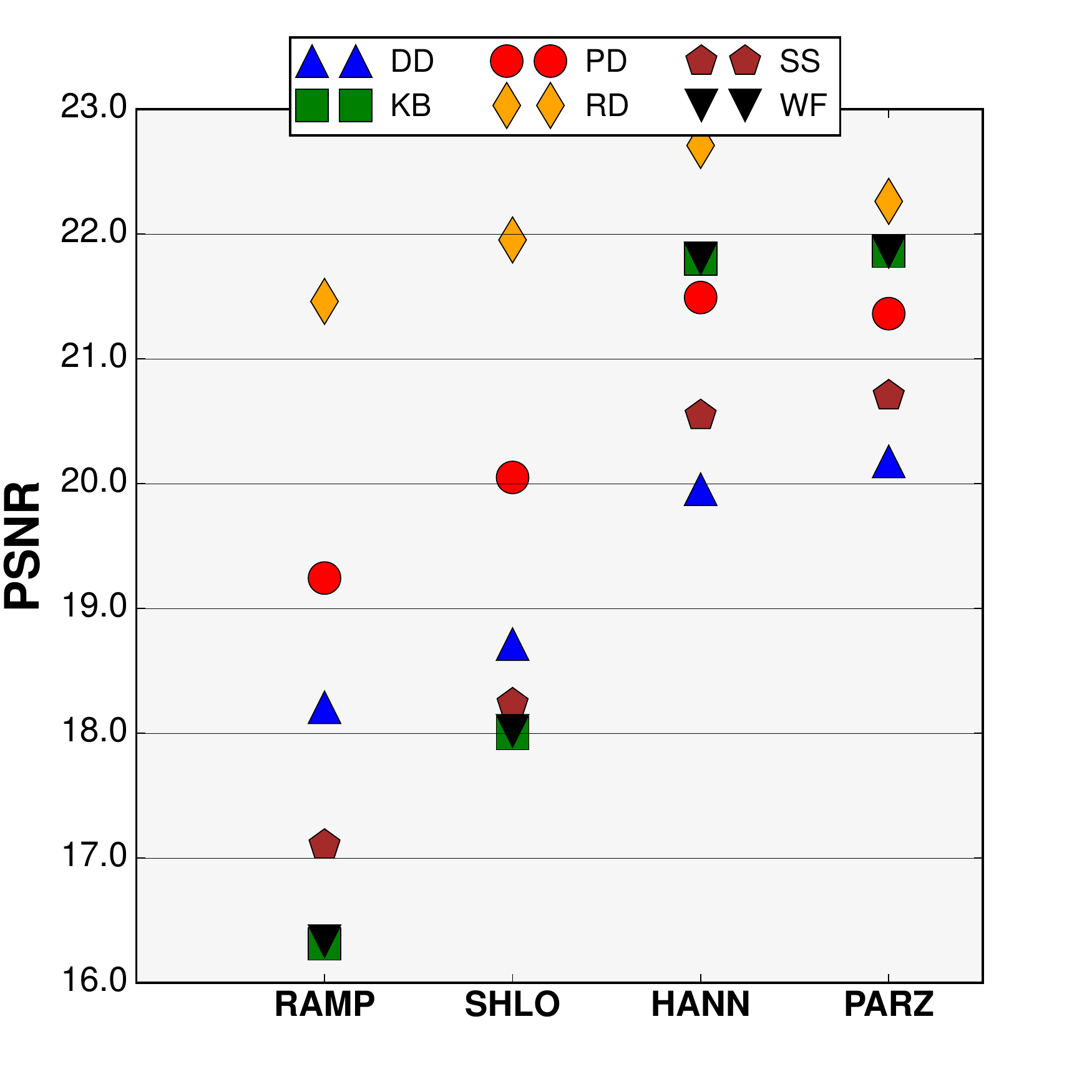}\label{anal-backproj-noise}  }}%
    \caption{Accuracy of the standalone backprojectors in performing FBP reconstruction with different filters (RAMP, SHLO, HANN, PARZ) of SL analytical sinograms.
             Reconstruction of (a) SL-FULL, (b) SL-UNDER and (c) SL-UCONSTR.}%
    \label{anal-backproj}%
\end{figure*}

The accuracy of the standalone forward projectors DD, KB, PD, RD, SS and WF with respect to SL-FULL is reported in
Tab.\ref{anal-forward}. The standalone backprojectors are used to perform FBP reconstructions with different filters
of SL-FULL, SL-UNDER and SL-UCONSTR\footnote{Results with SL-NOISE show the same trends
characterizing the reconstruction of SL-FULL, SL-UNDER and SL-UCONSTR and therefore are not shown.} and the corresponding results are illustrated in Fig.\ref{anal-backproj}.
The analysis in Tab.\ref{anal-forward} and Fig.\ref{anal-backproj} suggest two facts.
(i) The accuracy of the standalone projector
is not a good predictor of the accuracy of the standalone backprojector in analytical reconstruction: e.g., KB has the lowest 
PSNR value in Tab.\ref{anal-forward}, but it provides higher quality reconstruction of SL-FULL than PD and DD (Fig.\ref{anal-backproj-perfect}).
(ii) The performance of a backprojector is highly dependent on the characteristics of the dataset: e.g., SS has the highest PSNR score in  
Tab.\ref{anal-forward} and the best reconstruction quality for SL-FULL (Fig.\ref{anal-backproj-perfect}), but it performs poorly when reconstructing
underconstrained datasets (SL-UNDER and SL-UCONSTR in Fig.\ref{anal-backproj-under} and \ref{anal-backproj-noise}).

\begin{figure*}[!t]
  \centering
   \hspace*{0cm}\subfloat[FBP of DD sinogram]{{\includegraphics[width=6cm]{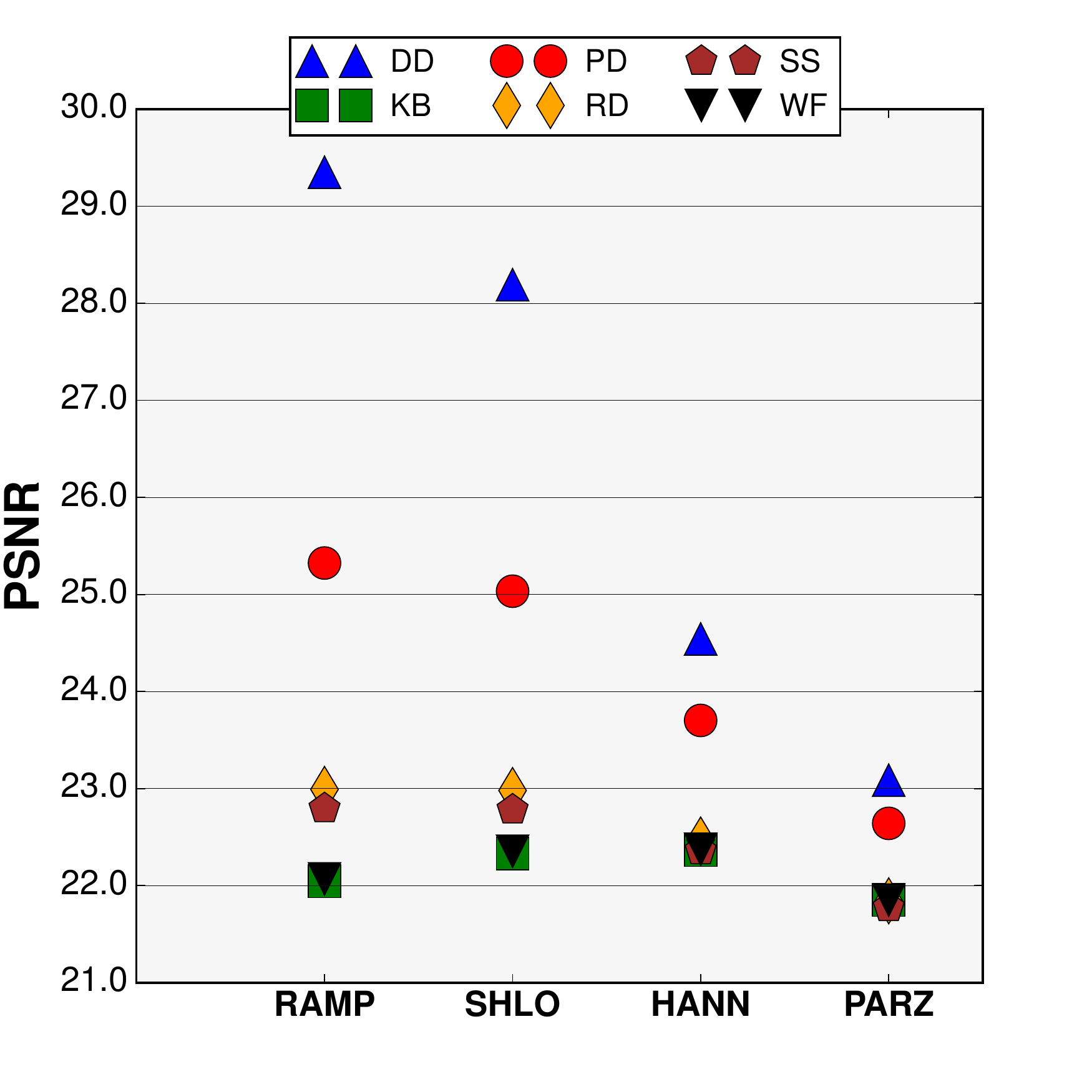}\label{anal-coupling-dd} }}%
   \hspace*{0.5cm}\subfloat[FBP of KB sinogram]{{\includegraphics[width=6cm]{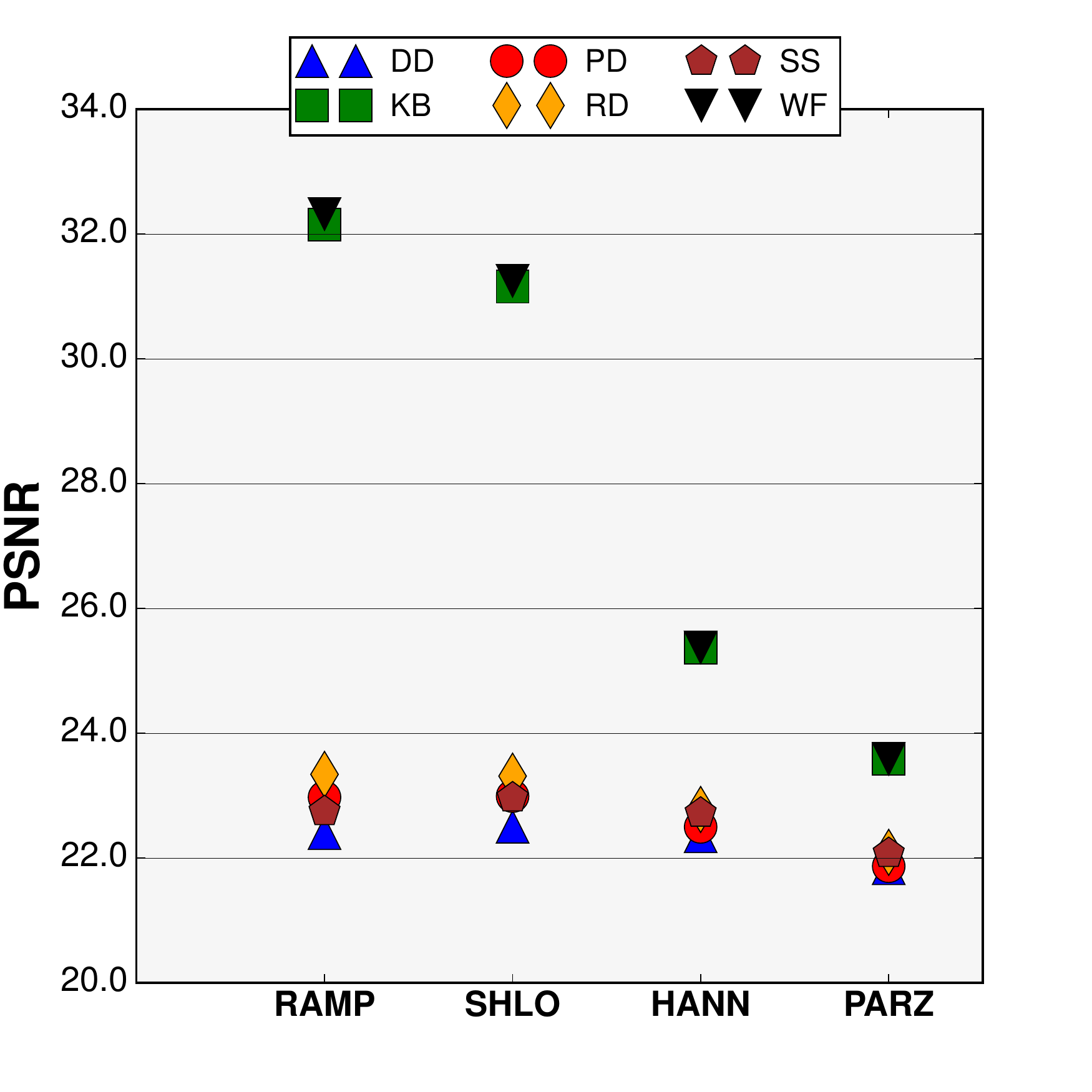}\label{anal-coupling-kb}  }}%
   \hspace*{0.5cm}\subfloat[FBP of PD sinogram]{{\includegraphics[width=6cm]{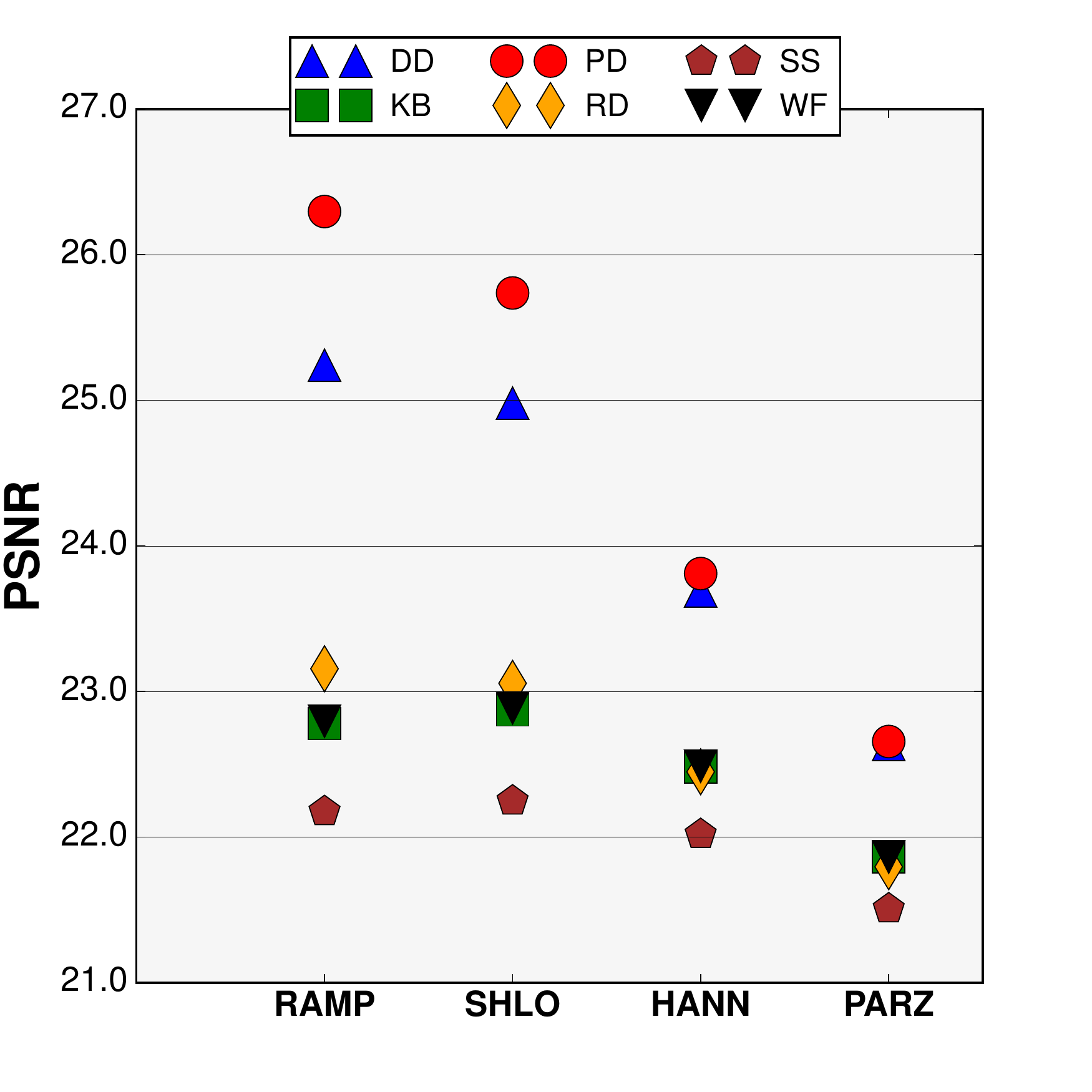}\label{anal-coupling-pd}  }}%
    \caption{FBP reconstructions of sinograms with 402 views $\times$ 256 pixels created by the DD, KB and PD forward projectors. 
             The reconstructions are perfomed with
             different filters by the DD, KB, PD, RD, SS and WF backprojectors.}%
    \label{anal-coupling}%
\end{figure*}

The experiment in Fig.\ref{anal-coupling} evaluates the reconstruction accuracy of well-sampled noiseless sinograms created by DD (Fig.\ref{anal-coupling-dd}),
KB (Fig.\ref{anal-coupling-kb}) and PD (Fig.\ref{anal-coupling-pd}). 
\begin{table}[H]\small
  \caption{Accuracy of the standalone forward projectors with respect to SL-FULL.}
  \label{anal-forward}
  \begin{center}
  \begin{tabular}{|c|c|c|c|c|c|c|}
\hline
       & DD & KB & PD & RD & SS & WF \\\hline
 PSNR  & 39.49 & 37.64 & 39.35 &  39.35 & 45.53 & 37.57 \\\hline
  \end{tabular}
\end{center}
\end{table}
\begin{figure*}[!t]
  \centering
   \hspace*{0.0cm}
   \subfloat[FBP of a DD undersampled sinogram]{{\includegraphics[width=6cm]{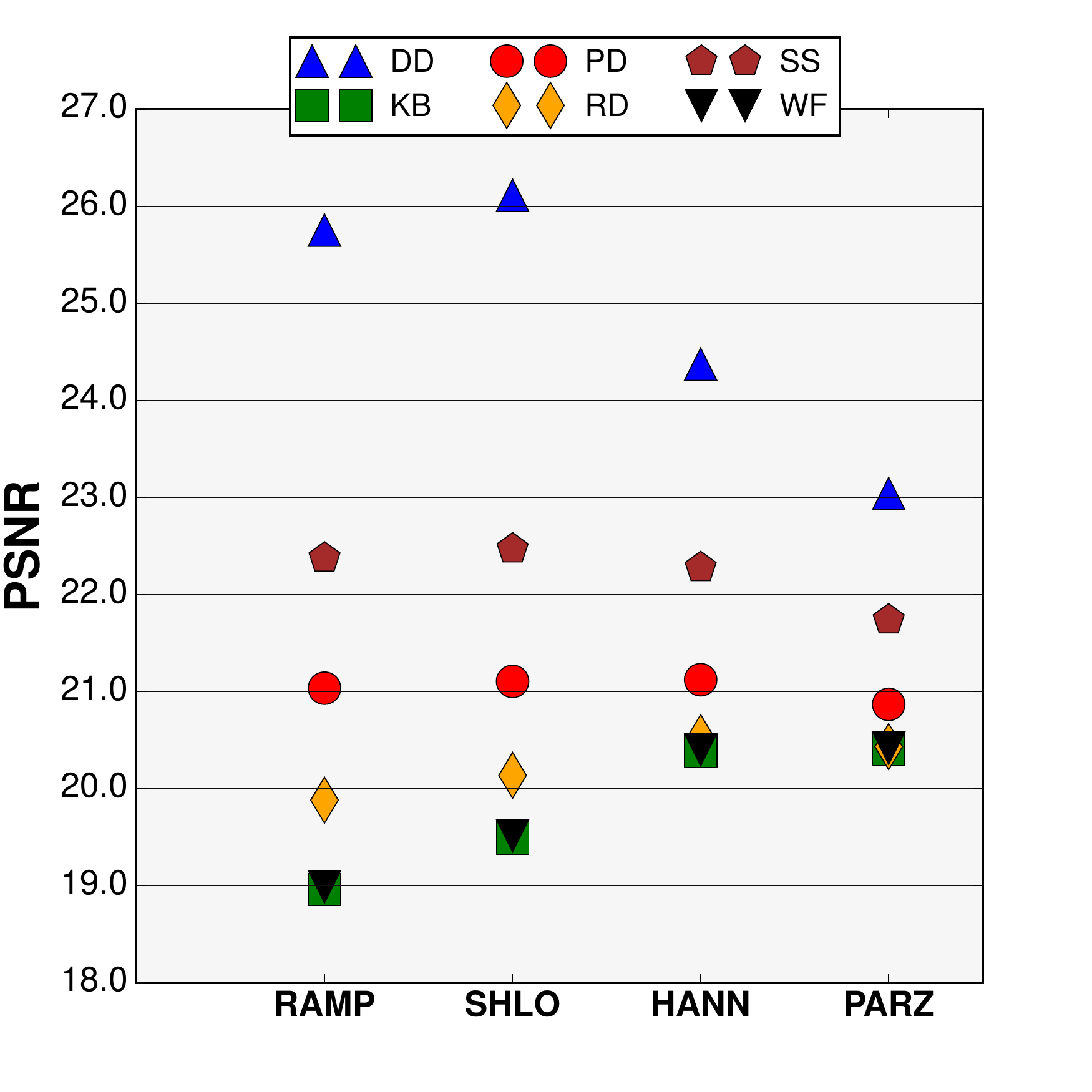}\label{anal-coupling-2-dd-under} }}%
   \hspace*{0.5cm}\subfloat[FBP of a KB noisy sinogram]{{\includegraphics[width=6cm]{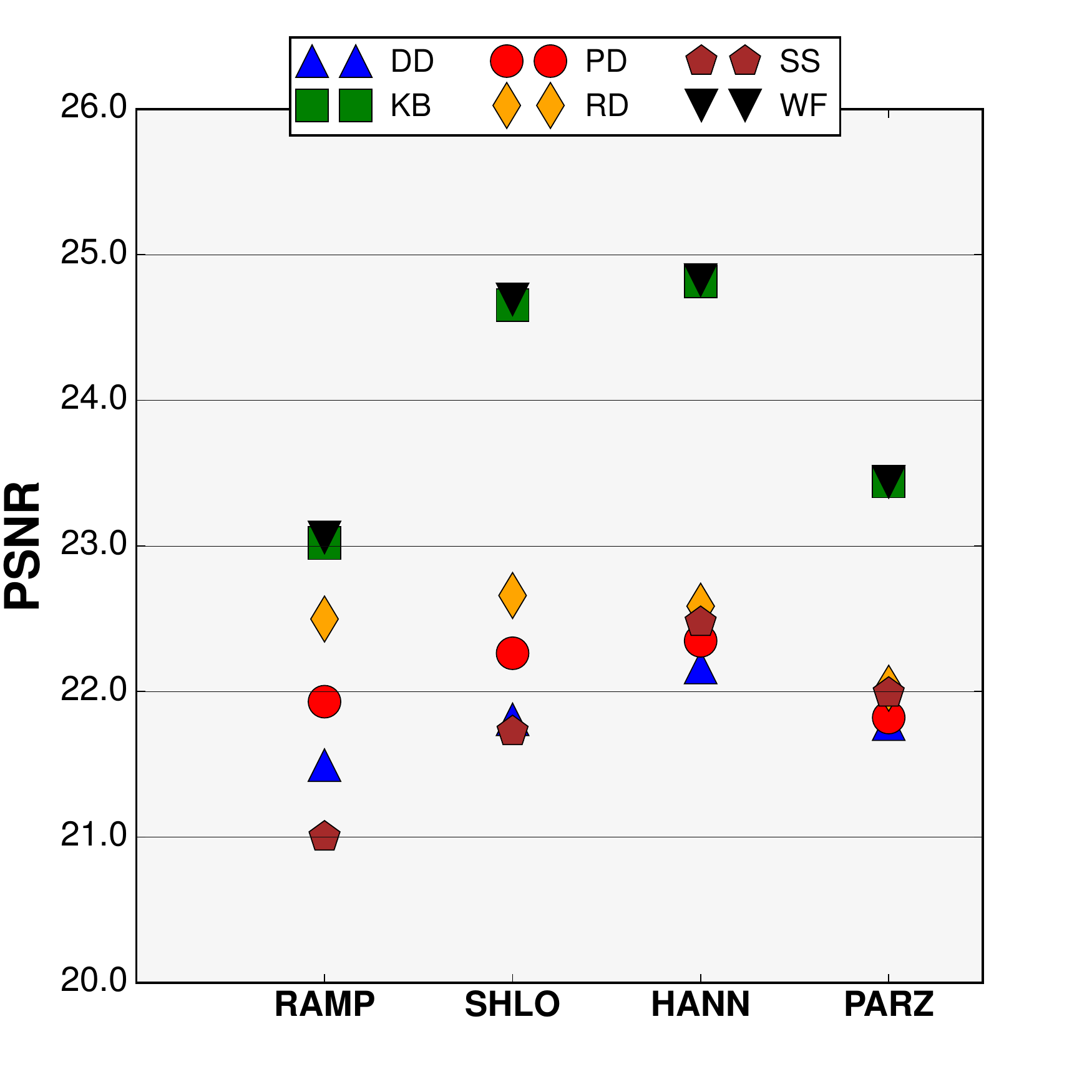}\label{anal-coupling-2-kb-noise} }}%
   \hspace*{0.5cm}\subfloat[FBP of a PD underconstrained sinogram]{{\includegraphics[width=6cm]{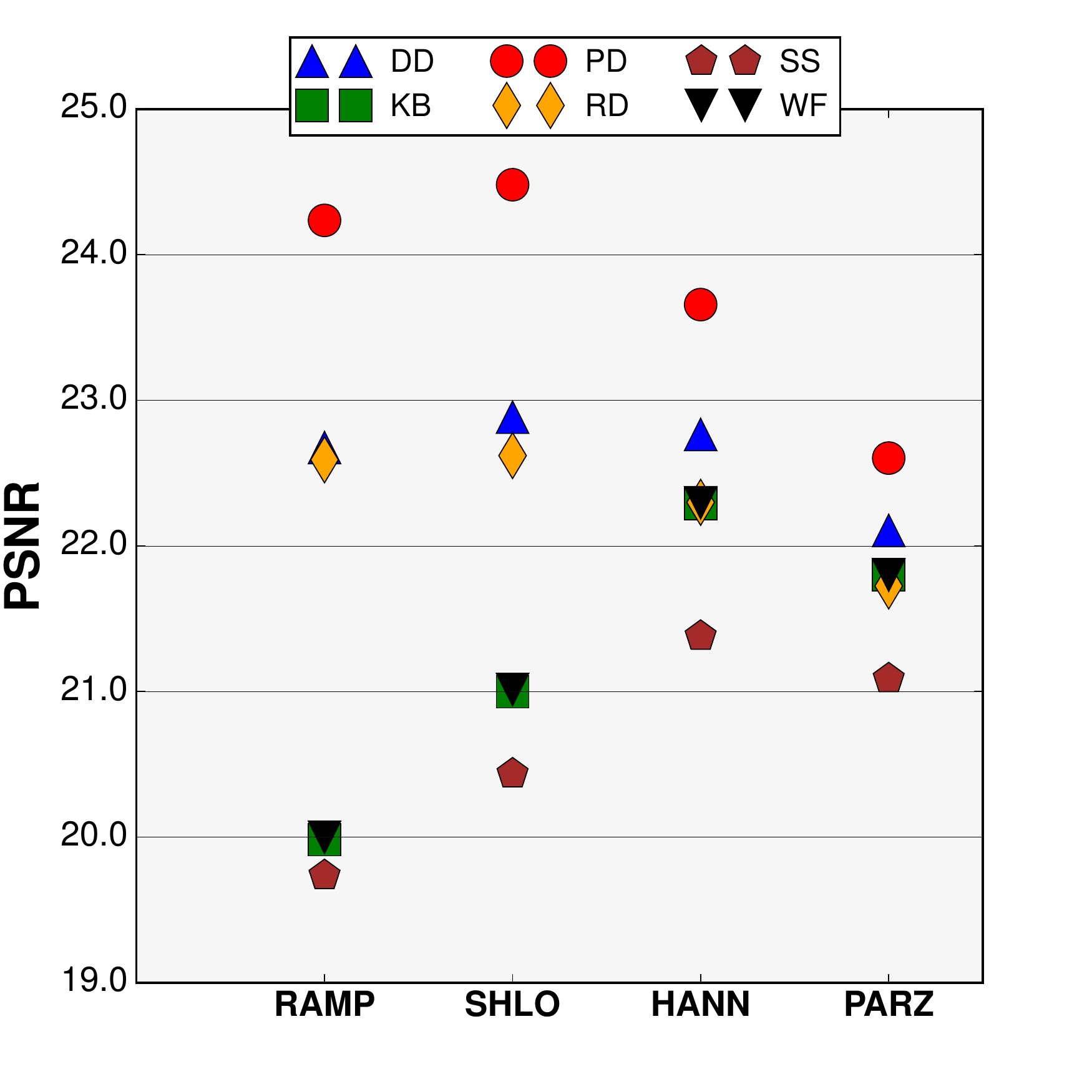}\label{anal-coupling-2-pd-under-noise}  }}%
   \caption{FBP reconstructions of (a) an undersampled sinogram with 100 views $\times$ 256 pixels created by the DD, 
            (b) a noisy sinogram with 402 views $\times$ 256 pixel and additional Poisson noise ($\sigma = 2\%$ of SL-FULL mean) created by KB and
            (c) an underconstrained sinogram with 100 views $\times$ 256 and additional Poisson noise ($\sigma = 2\%$ of SL-FULL mean)
            created by PD. The reconstructions are perfomed with
             different filters by the DD, KB, PD, RD, SS and WF backprojectors.}%
    \label{anal-coupling-2}%
\end{figure*}
\hspace{-0.6cm}The effect of the coupling projector-backprojector is clear: regardless of the filter choice,
the best reconstruction quality is achieved when the backprojector matches the operator used to compute the input sinogram. The weaker the action of the filter,
the more pronounced the impact of the coupling on the reconstruction accuracy.
The results of the FBP reconstructions in Fig.\ref{anal-coupling-2} show that the role of the coupling remains important even when dealing with undersampled
(Fig.\ref{anal-coupling-2-dd-under}), noisy (Fig.\ref{anal-coupling-2-kb-noise}) or
underconstrained (Fig.\ref{anal-coupling-2-pd-under-noise}) datasets. Considering that the performance of the standalone backprojectors can 
strongly vary as a function of the dataset (Fig.\ref{anal-backproj}), it is remarkable that undersampling and noise fail at breaking the effect of the 
coupling projector-backprojector. 
The important role of the coupling projector-backprojector is also clear when the sinograms are computed by RD, SS and WF (not shown).

\section{Operator coupling in iterative reconstruction}
\label{operator-coupling-iterative-reconstruction}
To study the coupling effect on the convergence of iterative algorithms, SL-FULL is reconstructed with
ADMM, PWLS, MLEM and SIRT. In each test, a different pair of forward and backward operators is used (Fig. 5-7).
Only results for selected combinations of tomographic operators are shown in this section for illustration.
The observed trends are however confirmed by all combinations.

\begin{figure*}[!t]
  \centering
   \hspace*{0.0cm}\subfloat[ADMM \textendash $\:\Radon$=RD]{{\includegraphics[width=8cm]{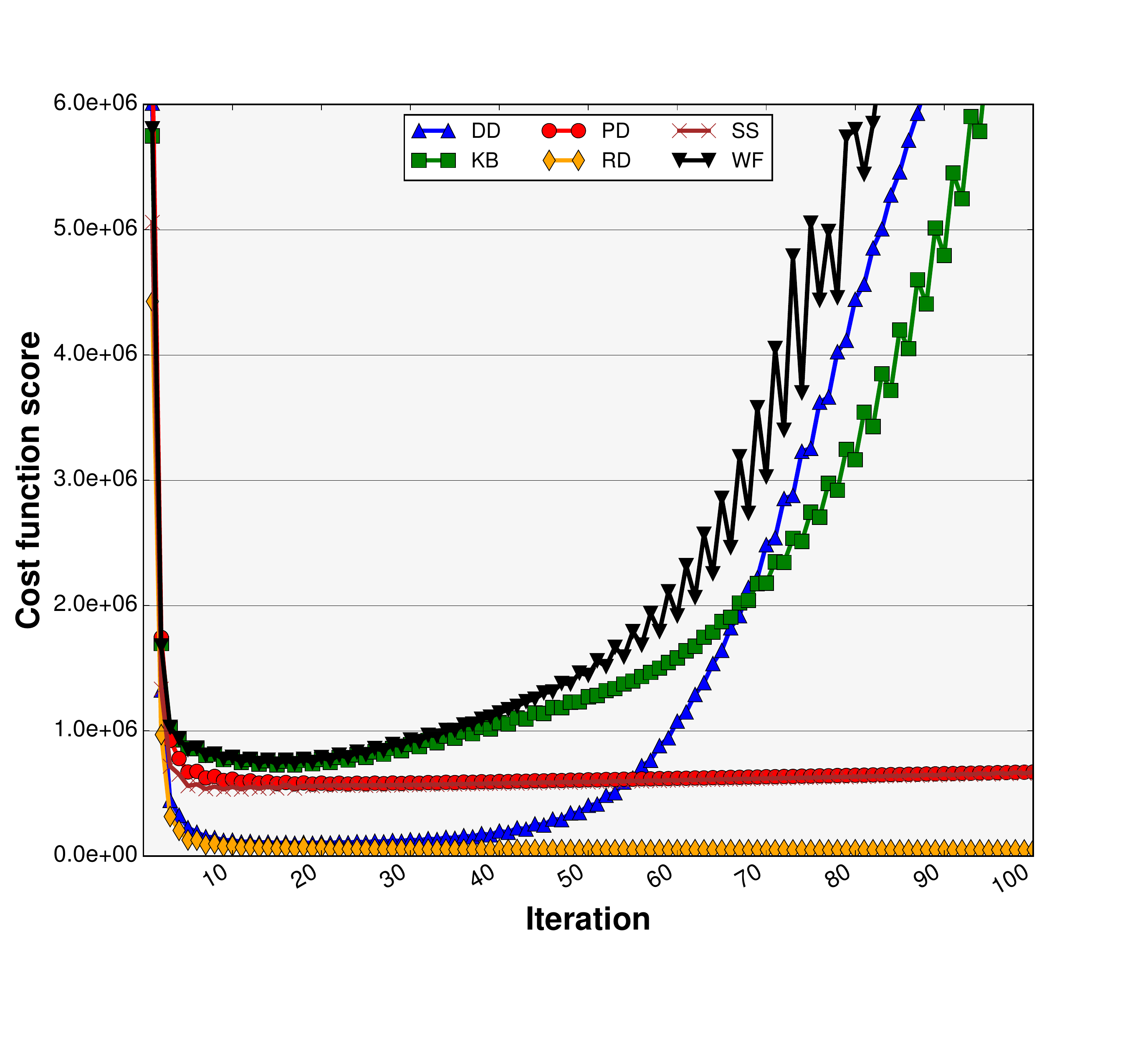}\label{iter-conv-admm-rd}  }}%
   \hspace*{1.0cm}\subfloat[ADMM \textendash $\:\Radon$=DD]{{\includegraphics[width=8cm]{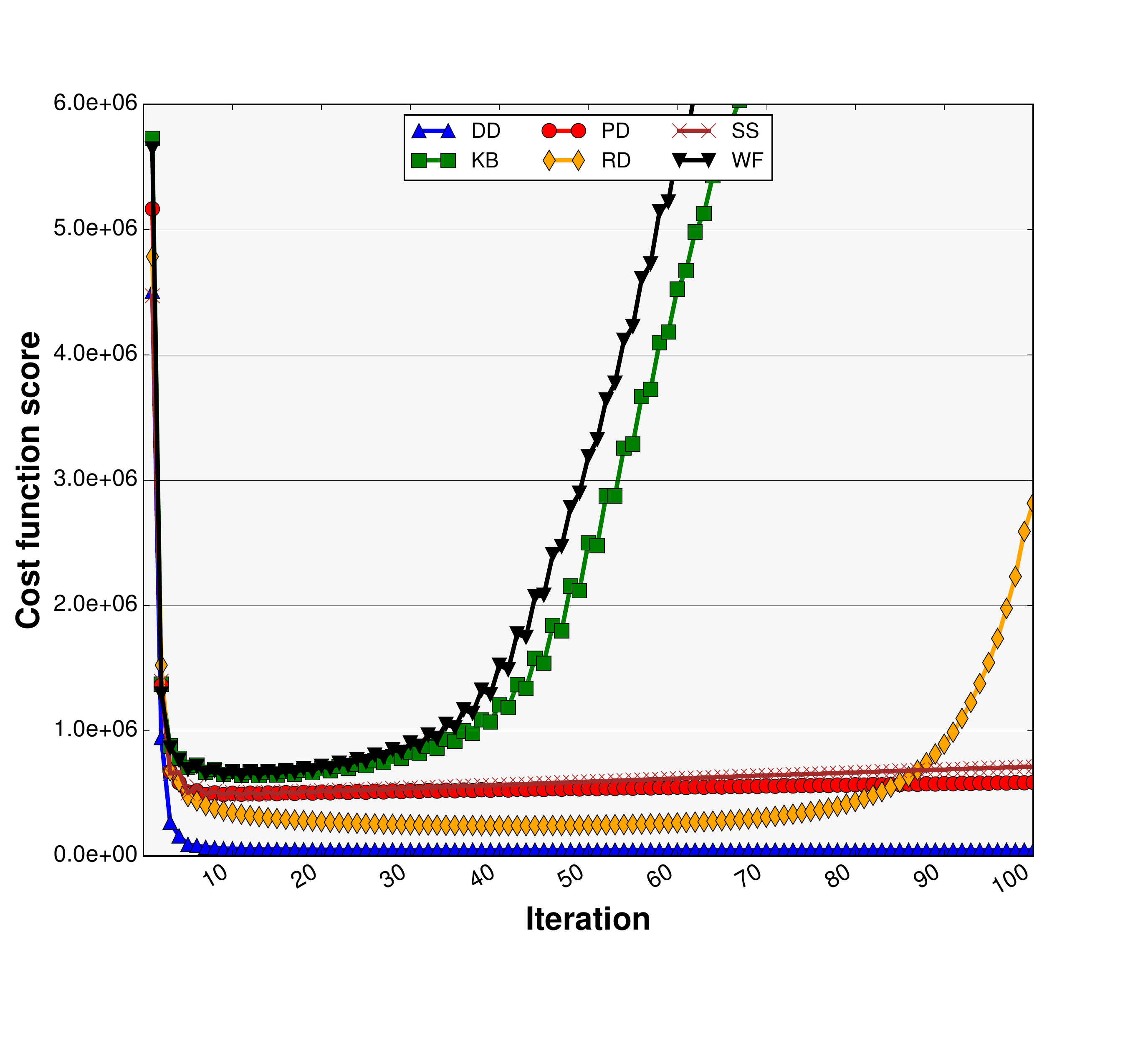}\label{iter-conv-admm-dd} }}%
    \caption{Study of convergence of the ADMM, using RD (Fig.\ref{iter-conv-admm-rd}) or DD (Fig.\ref{iter-conv-admm-dd}) as forward projectors
             combined to all six backprojectors considered in this study.}%
    \label{iter-conv-admm}%
\end{figure*}

ADMM converges and reaches the lowest value of the cost function when the backprojector matches the forward operator (Fig.\ref{iter-conv-admm}).
When the backprojector does not match the forward operator, three different scenarios are observed.
(i) ADMM converges but the cost function does not reach the minimum value (SS and PD curves in Fig.\ref{iter-conv-admm-rd}).
(ii) ADMM simply does not converge (KB and WF curves in Fig.\ref{iter-conv-admm-dd}). (iii) ADMM reaches the lowest value 
of the cost function before diverging (DD curve in Fig.\ref{iter-conv-admm-rd}).

\begin{figure*}[!t]
  \centering
   \hspace*{0.0cm}\subfloat[PWLS \textendash $\:\Radon$=WF]{{\includegraphics[width=8cm]{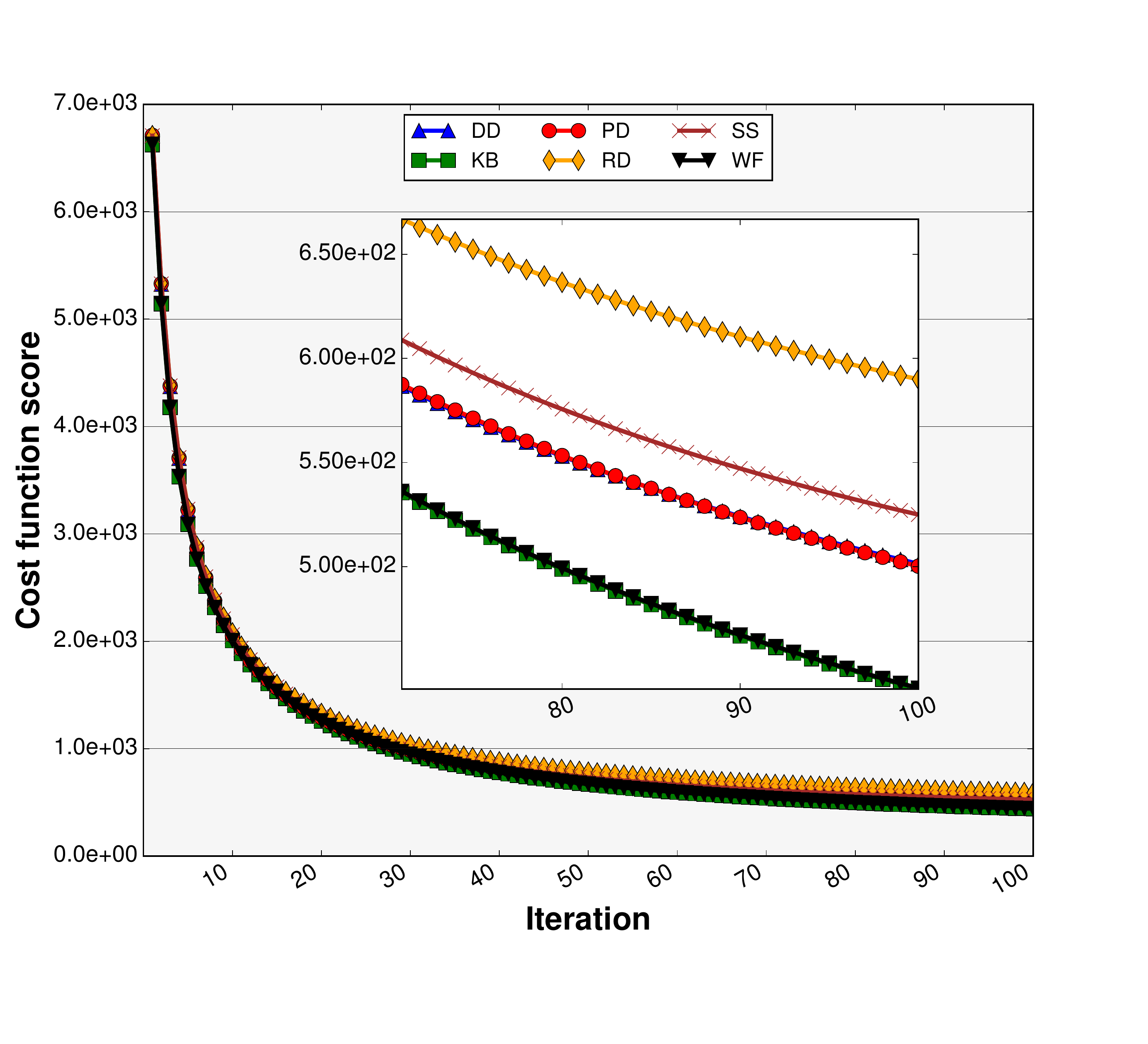}\label{iter-conv-sps-wf}  }}%
   \hspace*{1.0cm}\subfloat[PWLS \textendash $\:\Radon$=SS]{{\includegraphics[width=8cm]{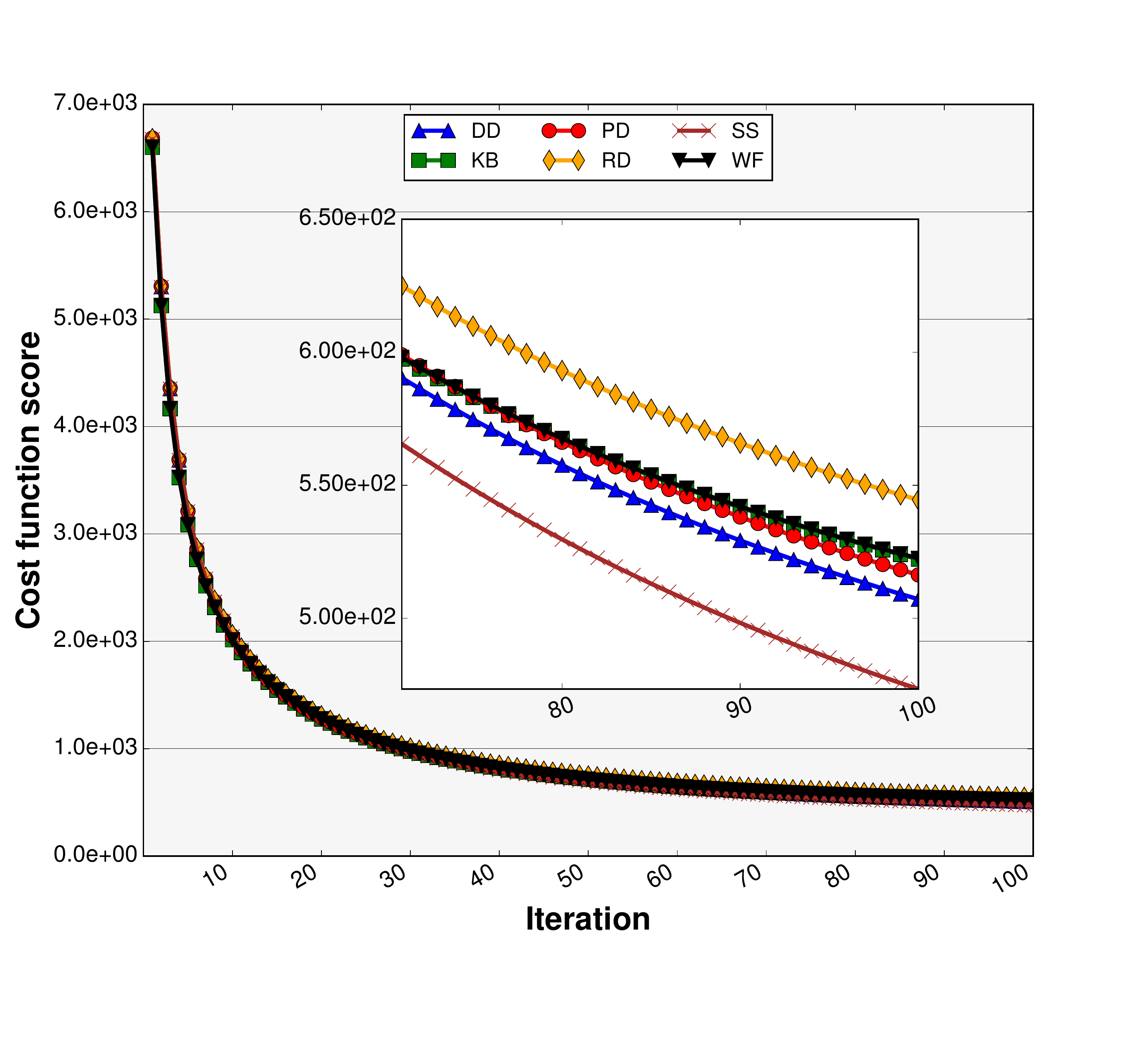}\label{iter-conv-sps-ss}  }}
    \caption{Study of convergence of the PWLS, using WF (Fig.\ref{iter-conv-sps-wf}) or SS (Fig.\ref{iter-conv-sps-ss}) as forward projectors
             combined to all six backprojectors considered in this study.}%
    \label{iter-conv-sps}%
\end{figure*}

Differently from ADMM, the convergence of PWLS is not endangered by a mismatch between tomographic operators.
Nevertheless, the cost function curve of PWLS with coupled operators is the lowest at each point after few initial iterations.
This is visible in the insets of Fig.\ref{iter-conv-sps-wf} and \ref{iter-conv-sps-ss}.

\begin{figure*}[!t]
  \centering
   \hspace*{0.0cm}\subfloat[MLEM \textendash $\:\Radon$=PD]{{\includegraphics[width=8cm]{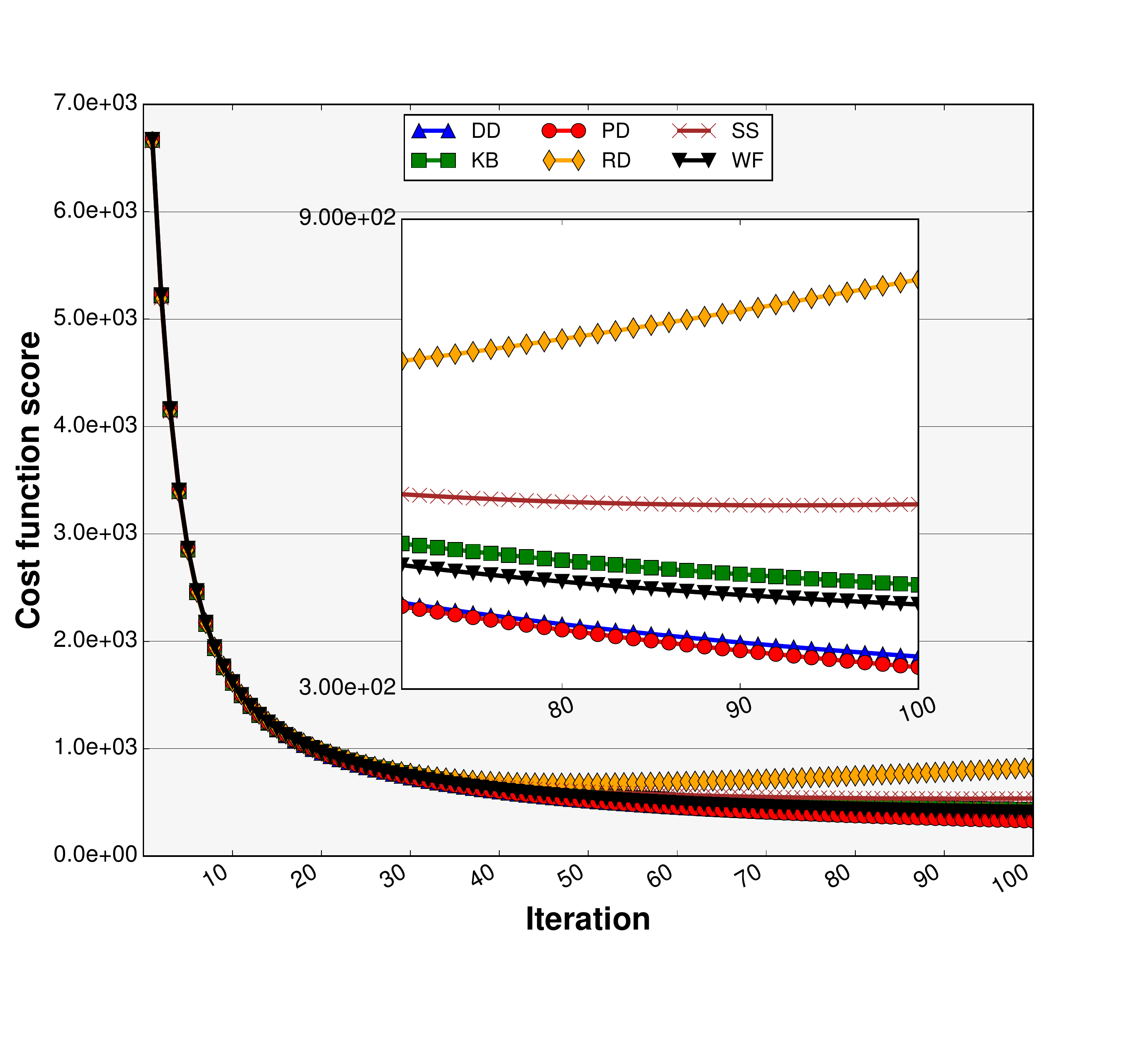}\label{iter-conv-em-pd}  }}%
   \hspace*{1.0cm}\subfloat[SIRT \textendash $\:\Radon$=PD]{{\includegraphics[width=8cm]{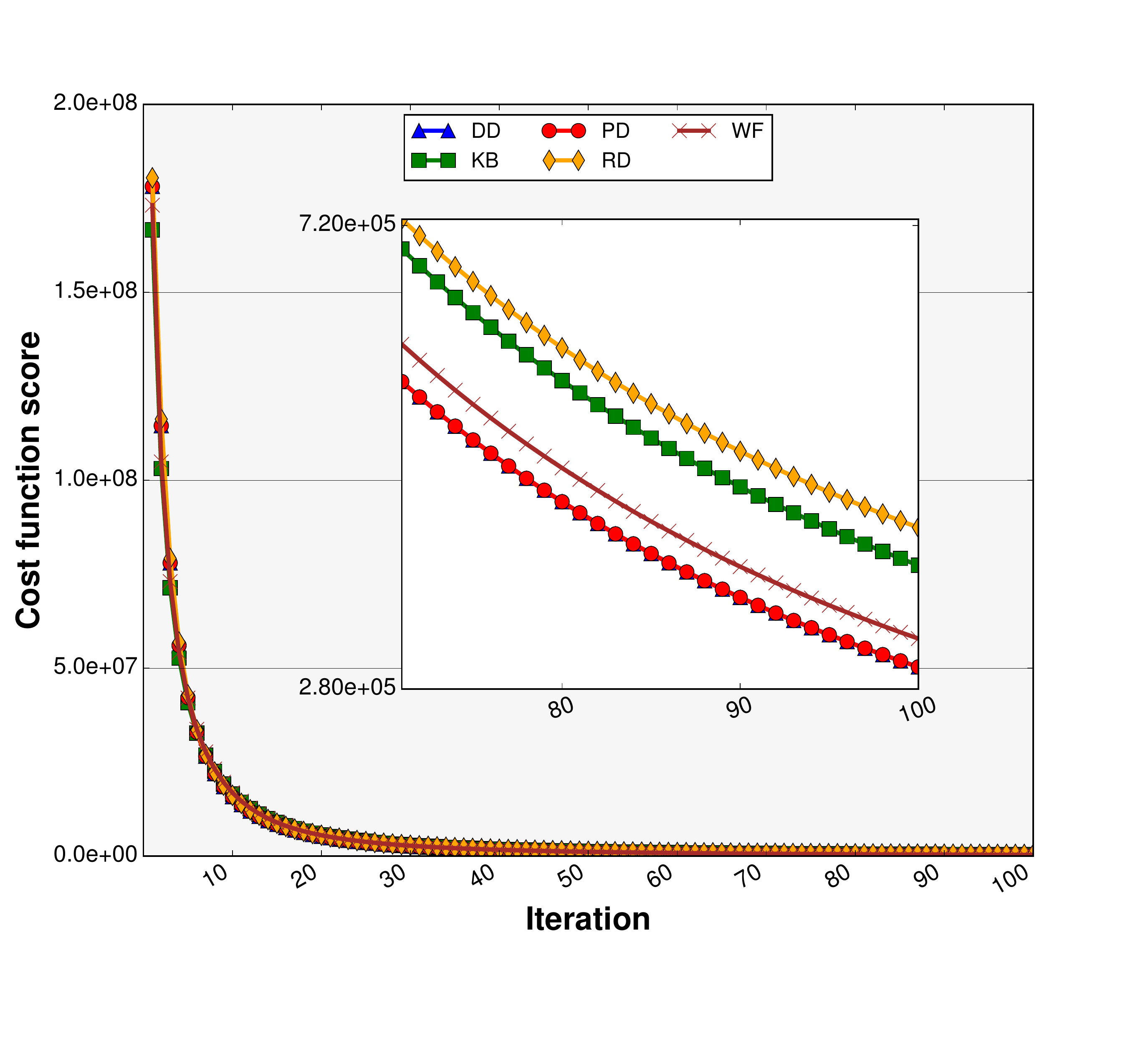}\label{iter-conv-sirt-pd}  }}
    \caption{Study of convergence of the MLEM and SIRT, both using PD as forward projector
             combined to all six backprojectors considered in this study.}%
    \label{iter-conv-em-sirt}%
\end{figure*}

MLEM and SIRT behave similarly to PWLS: the matching between forward projector and backprojector is not essential
to guarantee convergence, but is required to obtain the lowest cost function curve at each point, as shown in 
the insets of Fig.\ref{iter-conv-em-pd} and \ref{iter-conv-sirt-pd}. Despite this similarity to PWLS, MLEM and SIRT can, instead, easily
``explode'' with an undersampled or noisy dataset if the operators are not coupled. For this reason, no reconstruction of
underconstrained datasets done by SIRT and only few cases with MLEM are shown in the following. 
SIRT and MLEM share a common aspect: the computation of the diagonal matrix $\mathbf{C} = \{ c_{jj} = 1/\sum_{i}a_{ij}\}$ is necessary, where
$\{a_{ij}\}$ are the elements of the matrix representation of $\Radon$. The $c_{jj}$'s can be efficiently calculated as
$\Radon^{*}(\mathds{1})\,, \,\mathbb{R}^{M\times N} \ni \mathds{1} = \{(\mathds{1})_{ij} = 1 \,\,\,\forall\:i,j\}$.   
This computation can be rather sensitive and produce very high values at the image boudaries, compromising the stability of the iterative
procedure especially when using uncoupled projectors.
On the other hand, since
ADMM and PWLS do not involve potentially sensitive computations, tests of these algorithms were not restricted to specific datasets or projector pairs.

The results in Fig.\ref{iter-conv-admm}, \ref{iter-conv-sps} and \ref{iter-conv-em-sirt} clearly illustrate the influence of the
coupling projector-backprojector on the convergence of all considered iterative 
procedures: the best performance is achieved only when the operators match.
The level of the cost function of an iterative algorithm after a certain amount of iterations is not completely related to the reconstruction accuracy, or, in other words,
reaching the minimum of the cost function does not necessarily mean reaching the closest possible approximation to the original phantom.
Additional experiments focusing on the reconstruction accuracy have been performed. Reconstructions are displayed when differences
can be perceived at visual inspection.

\begin{figure*}[!t]
  \centering
   \hspace*{0.0cm}\subfloat[$R^{*}$=SS \textendash\: PSNR=20.13]{{\includegraphics[width=4.5cm]{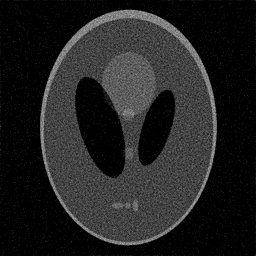}\label{iter-admm-noise-ss-ss} }}%
   \hspace*{0.0cm}\subfloat[$R^{*}$=DD \textendash\: PSNR=17.15]{{\includegraphics[width=4.5cm]{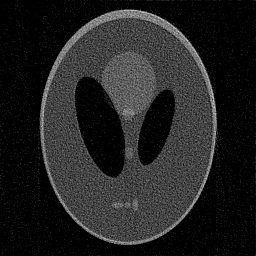} \label{iter-admm-noise-ss-dd} }}%
   \hspace*{0.0cm}\subfloat[$R^{*}$=KB \textendash\: PSNR=16.50]{{\includegraphics[width=4.5cm]{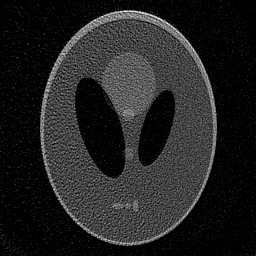}\label{iter-admm-noise-ss-kb} }}%
   \hspace*{0.0cm}\subfloat[$R^{*}$=RD \textendash\: PSNR=18.51]{{\includegraphics[width=4.5cm]{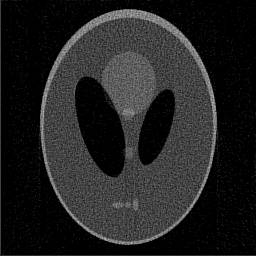} \label{iter-admm-noise-ss-rd} }}\\      
    \caption{ADMM reconstructions of SL-NOISE using SS as forward projector and SS, DD, KB, RD as backprojectors.}%
    \label{iter-admm-noise}%
\end{figure*}

\begin{figure*}[!t]
  \centering
   \hspace*{0.0cm}\subfloat[$R^{*}$=KB \textendash\: PSNR=16.87]{{\includegraphics[width=4.5cm]{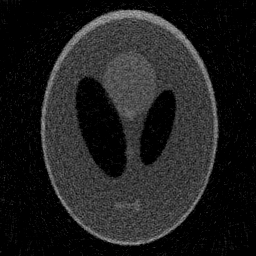}\label{iter-admm-under-noise-kb-kb} }}%
   \hspace*{0.0cm}\subfloat[$R^{*}$=PD \textendash\: PSNR=12.53]{{\includegraphics[width=4.5cm]{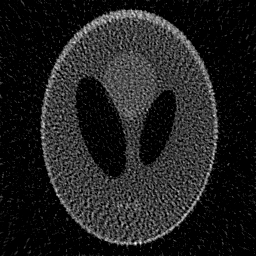} \label{iter-admm-under-noise-kb-pd} }}%
   \hspace*{0.0cm}\subfloat[$R^{*}$=RD \textendash\: PSNR=14.17]{{\includegraphics[width=4.5cm]{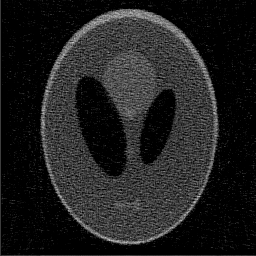}\label{iter-admm-under-noise-kb-rd} }}%
   \hspace*{0.0cm}\subfloat[$R^{*}$=WF \textendash\: PSNR=16.86]{{\includegraphics[width=4.5cm]{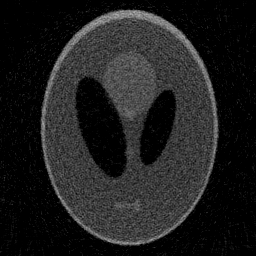} \label{iter-admm-under-noise-kb-wf} }}\\      
    \caption{ADMM reconstructions of SL-UCONSTR using KB as forward projector and KB, PD, RD, WF as backprojectors.}%
    \label{iter-admm-under-noise}%
\end{figure*}

Table \ref{iter-admm-under} presents the results of ADMM reconstructions of SL-UNDER with the PD forward projector.
The best quality is achieved when the PD backprojector is used. Nevertheless, differences are relatively small and the reconstructions
look very similar. The coupling has a much stronger effect when reconstructing SL-NOISE, as shown in Fig.\ref{iter-admm-noise}:
the best ADMM reconstruction is obtained when the operators match (SS, in this case) and differences in PSNR are up to 3.6 dB.
At visual inspection, reconstructions in Fig.\ref{iter-admm-noise-ss-dd}, \ref{iter-admm-noise-ss-kb} and \ref{iter-admm-noise-ss-rd} are
slightly more degraded than in Fig.\ref{iter-admm-noise-ss-ss}, as suggested by the PSNR score.
Results in Fig.\ref{iter-admm-under-noise} show once again the great impact 
of the coupling effect on the reconstruction accuracy in presence of noise.
Since KB and WF are both based on the gridding method and are highly coupled 
(as also resulting from the previous analysis), the reconstruction in Fig.\ref{iter-admm-under-noise-kb-wf}
is nearly identical to the one performed with matching operators in Fig.\ref{iter-admm-under-noise-kb-kb}.
The combination of a noisy underconstrained dataset and poorly coupled operators leads, instead, to strongly degraded ADMM reconstructions
(Fig.\ref{iter-admm-under-noise-kb-pd} and Fig.\ref{iter-admm-under-noise-kb-rd}).

\begin{figure*}[!t]
  \centering
   \hspace*{0.0cm}\subfloat[$R^{*}$=KB \textendash\: PSNR=19.90]{{\includegraphics[width=4.5cm]{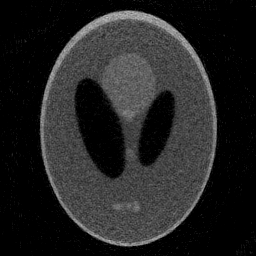}\label{iter-sps-under-noise-kb-kb} }}%
   \hspace*{0.0cm}\subfloat[$R^{*}$=SS \textendash\: PSNR=17.84]{{\includegraphics[width=4.5cm]{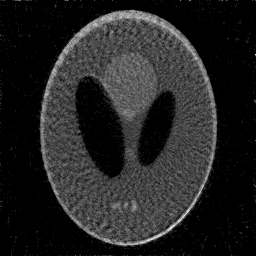} \label{iter-sps-under-noise-kb-ss} }}%
   \hspace*{0.0cm}\subfloat[$R^{*}$=DD \textendash\: PSNR=19.61]{{\includegraphics[width=4.5cm]{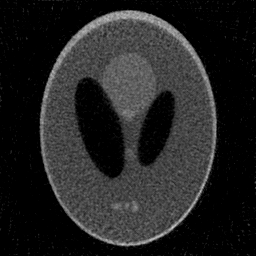}\label{iter-sps-under-noise-kb-dd} }}%
   \hspace*{0.0cm}\subfloat[$R^{*}$=PD \textendash\: PSNR=19.62]{{\includegraphics[width=4.5cm]{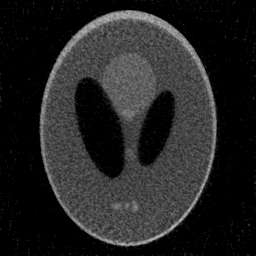} \label{iter-sps-under-noise-kb-pd} }}\\      
    \caption{PWLS reconstructions of SL-UCONSTR using KB as forward projector and KB, SS, DD, PD as backprojectors.}%
    \label{iter-sps-under-noise}%
\end{figure*}

The PSNR values in Tab.\ref{iter-sps-under}(a) and \ref{iter-sps-under}(b) correspond, respectively, to PWLS reconstruction of SL-UNDER using the KB forward projector 
with KB, DD, SS and WF backprojectors and of SL-NOISE using the SS forward projector  with SS, DD, WF and PD backprojectors. 
For PWLS, the coupling projector-backprojector has slightly more impact in presence of undersampled data than of purely noisy data:
the spread of PSNR values in Tab.\ref{iter-sps-under}(a) is, indeed, a bit larger than for the values in Tab.\ref{iter-sps-under}(b).
Similarly to the results of Fig.\ref{iter-admm-under-noise}, the PWLS reconstruction of underconstrained datasets with coupled projectors
has the highest accuracy (Fig.\ref{iter-sps-under-noise-kb-kb}), whereas severe artifacts can occur when reconstructing an underconstrained
dataset with uncoupled operators
(Fig.\ref{iter-sps-under-noise-kb-ss}).

\begin{figure*}[!t]
  \centering
   \hspace*{0.0cm}\subfloat[$R^{*}$=PD \textendash\: PSNR=20.74]{{\includegraphics[width=4.5cm]{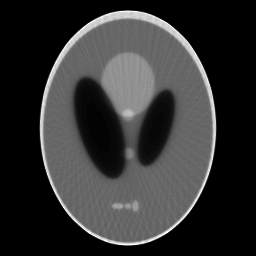}\label{iter-em-under-pd-pd} }}%
   \hspace*{0.0cm}\subfloat[$R^{*}$=RD \textendash\: PSNR=10.63]{{\includegraphics[width=4.5cm]{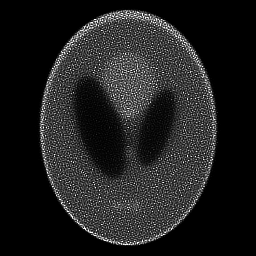} \label{iter-em-under-pd-rd} }}%
   \hspace*{0.0cm}\subfloat[$R^{*}$=KB \textendash\: PSNR=19.42]{{\includegraphics[width=4.5cm]{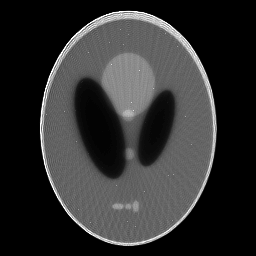}\label{iter-em-under-pd-kb} }}%
    \caption{MLEM reconstructions of SL-UNDER using PD as forward projector and PD, RD, KB as backprojectors.}%
    \label{iter-em-under-noise}%
\end{figure*}

Reconstructions with MLEM and SIRT are very sensitive to the coupling effect with both undersampled and noisy datasets.
Several reconstruction attempts for SL-UNDER, SL-NOISE and SL-UCONSTR using these algorithms with non-matching operators 
failed, as the procedure quickly diverges after
few iterations.
Figure \ref{iter-em-under-noise} shows an experiment with MLEM, PD forward projector
and PD, RD and KB backprojectors: the reconstruction with coupled operators (Fig.\ref{iter-em-under-pd-pd}) is once again characterized
by the highest accuracy.
\begin{table}[!t]\small
  \caption{PSNR scores of ADMM reconstructions of SL-UNDER using PD as forward projector and PD, KB, RD, WF as backprojectors.}
  \label{iter-admm-under}
  \begin{center}
  \begin{tabular}{|c|c|c|c|c|c|c|}
\hline
       & $R^{*}$=PD & $R^{*}$=KB & $R^{*}$=RD & $R^{*}$=WF \\\hline
 PSNR  & 22.06 & 21.67 & 21.29 & 21.46 \\\hline
  \end{tabular}
\end{center}
\end{table}
\hspace{-0.6cm}The last experiment is designed to roughly estimate the impact of the coupling projector-backprojector
on the reconstruction quality with respect to other two fundamental components: 
physical constraints (i.e., setting to zero all negative pixels at each
iteration) and optimal number of iterations. As example we show here the results for SL-UCONSTR and the ADMM.
The highest PSNR in Tab.\ref{admm-parts} corresponds to case (1), where all three components (coupling, constraints, optimal number of iterations)
are present. The interesting result is that case (2), that relies only on coupled operators, achieves a better reconstruction quality than
case (3), where constraints and optimal number of iterations are kept, but the operators are not matching.
This experiment gives a hint of the fact that, in some cases, the coupling projector-backprojector could even play a more decisive role than other
crucial factors on the accuracy of an iterative algorithm. 
To validate the generality of these last results, further in-depth analysis is required, subject of future work.
\begin{table}[!t]\small
  \centering
  \caption{PSNR scores of PSWS reconstructions of SL-UNDER using PD as forward projector (left) and SL-NOISE 
           using SS as forward projector (right).}
  \label{iter-sps-under}
  \begin{tabular}{|c|c|c|c|c|c|c|}
\hline
       & $R^{*}$=KB & $R^{*}$=DD & $R^{*}$=SS & $R^{*}$=WF \\\hline
 PSNR  & 22.09 & 19.15 & 19.24 & 21.50 \\\hline
  \end{tabular}\\[1.5em]
  \begin{tabular}{|c|c|c|c|c|c|c|}
\hline
       & $R^{*}$=SS & $R^{*}$=DD & $R^{*}$=WF & $R^{*}$=PD \\\hline
 PSNR  & 23.04 & 22.51 & 22.39 & 22.47 \\\hline
  \end{tabular}
\end{table}
\begin{table}[!t]\small
  \caption{Three different ADMM reconstructions of SL-UCONSTR. Case (1): coupled operators + constraints + optimal number of iterations.
           Case (2): coupled operators. Case (3): constraints + optimal number of iterations.}
  \label{admm-parts}
  \begin{center}
  \begin{tabular}{|c|c|c|c|c|c|c|}
\hline
       & Case 1 & Case 2 & Case 3 \\\hline
 PSNR  & 19.69 & 18.97 & 18.10 \\\hline
  \end{tabular}
\end{center}
\end{table}

\section{Conclusion}
This work is an experimental study on the impact of the coupling projector-backprojector in iterative reconstruction schemes.
Since iterative algorithms call the tomographic operators few times per iteration, it can be expected
that the level of matching between the actual implementation of the forward projector and backprojector can deeply
affect the performance of the entire iterative procedure.

A framework consisting of four iterative methods (the alternate direction method of multipliers, the 
penalized weighted least squares, the maximum-likelihood expectation maximization and the simultaneous iterative
algebraic technique) working with six different projectors (distance-driven, pixel-driven, ray-driven, slant-stacking 
and two gridding methods) has been conceived to test the aforementioned hypothesis.

All iterative experiments on simulated data clearly show that 
the performance of every selected method is deeply affected by the coupling projector-backprojector
in terms of convergence and accuracy. The best convergence behaviour and the highest reconstruction quality
are systematically obtained when the tomographic operators match.
This conclusion holds regardless of the nature of the input tomographic dataset in terms of angular sampling
or SNR.
Moreover, there is indication that the coupling projector-backprojector may represent one of the major players
determining the performance of an iterative algorithm, even with respect to physical constraints or optimal number of iterations.

The results of this study indicate that it would be strongly advisable for users and developers of software packages for iterative
tomographic reconstructions to always select projector pairs with a high mathematical affinity and to carefully assess and validate
the degree of coupling of the used implementations. This strategy is important
to avoid results systematically characterized by suboptimal accuracy.

\bibliographystyle{IEEEtran}
\bibliography{coupling_forward_adjoint}

\begin{thebibliography}{10}
\providecommand{\url}[1]{#1}
\csname url@samestyle\endcsname
\providecommand{\newblock}{\relax}
\providecommand{\bibinfo}[2]{#2}
\providecommand{\BIBentrySTDinterwordspacing}{\spaceskip=0pt\relax}
\providecommand{\BIBentryALTinterwordstretchfactor}{4}
\providecommand{\BIBentryALTinterwordspacing}{\spaceskip=\fontdimen2\font plus
\BIBentryALTinterwordstretchfactor\fontdimen3\font minus
  \fontdimen4\font\relax}
\providecommand{\BIBforeignlanguage}[2]{{%
\expandafter\ifx\csname l@#1\endcsname\relax
\typeout{** WARNING: IEEEtran.bst: No hyphenation pattern has been}%
\typeout{** loaded for the language `#1'. Using the pattern for}%
\typeout{** the default language instead.}%
\else
\language=\csname l@#1\endcsname
\fi
#2}}
\providecommand{\BIBdecl}{\relax}
\BIBdecl

\bibitem{Hounsfield1973}
\BIBentryALTinterwordspacing
G.~N. Hounsfield, ``Computerized transverse axial scanning (tomography): Part
  1. description of system,'' \emph{The British Journal of Radiology}, vol.~46,
  no. 552, pp. 1016--1022, dec 1973. [Online]. Available:
  \url{http://dx.doi.org/10.1259/0007-1285-46-552-1016}
\BIBentrySTDinterwordspacing

\bibitem{Herman2009}
G.~T. Herman, \emph{Fundamentals of Computerized Tomography: Image
  Reconstruction from Projections (Advances in Computer Vision and Pattern
  Recognition)}.\hskip 1em plus 0.5em minus 0.4em\relax Springer, 2009.

\bibitem{Herman1973}
\BIBentryALTinterwordspacing
G.~T. Herman, A.~Lent, and S.~W. Rowland, ``{ART}: Mathematics and
  applications,'' \emph{Journal of Theoretical Biology}, vol.~42, no.~1, pp.
  1--32, nov 1973. [Online]. Available:
  \url{http://dx.doi.org/10.1016/0022-5193(73)90145-8}
\BIBentrySTDinterwordspacing

\bibitem{Gilbert1972}
\BIBentryALTinterwordspacing
P.~Gilbert, ``Iterative methods for the three-dimensional reconstruction of an
  object from projections,'' \emph{Journal of Theoretical Biology}, vol.~36,
  no.~1, pp. 105--117, jul 1972. [Online]. Available:
  \url{http://dx.doi.org/10.1016/0022-5193(72)90180-4}
\BIBentrySTDinterwordspacing

\bibitem{Andersen1984}
\BIBentryALTinterwordspacing
A.~H. Andersen and A.~C. Kak, ``Simultaneous algebraic reconstruction technique
  ({SART}): A superior implementation of the art algorithm,'' \emph{Ultrasonic
  Imaging}, vol.~6, no.~1, pp. 81--94, jan 1984. [Online]. Available:
  \url{http://dx.doi.org/10.1177/016173468400600107}
\BIBentrySTDinterwordspacing

\bibitem{Kaczmarz1937}
S.~Kaczmarz, ``{Angen\"{a}herte Aufl\"{o}sung von Systemen linearer
  Gleichungen},'' \emph{Bulletin International de l'Acad\'{e}mie Polonaise des
  Sciences et des Lettres}, vol.~35, pp. 355--357, 1937.

\bibitem{Shepp1982}
\BIBentryALTinterwordspacing
L.~A. Shepp and Y.~Vardi, ``Maximum likelihood reconstruction for emission
  tomography,'' \emph{{IEEE} Transactions on Medical Imaging}, vol.~1, no.~2,
  pp. 113--122, oct 1982. [Online]. Available:
  \url{http://dx.doi.org/10.1109/TMI.1982.4307558}
\BIBentrySTDinterwordspacing

\bibitem{Erdogan1999}
\BIBentryALTinterwordspacing
H.~Erdogan and J.~A. Fessler, ``Ordered subsets algorithms for transmission
  tomography,'' \emph{Physics in Medicine and Biology}, vol.~44, no.~11, p.
  2835, 1999. [Online]. Available:
  \url{http://stacks.iop.org/0031-9155/44/i=11/a=311}
\BIBentrySTDinterwordspacing

\bibitem{Fessler1994}
\BIBentryALTinterwordspacing
J.~Fessler, ``Penalized weighted least-squares image reconstruction for
  positron emission tomography,'' \emph{{IEEE} Transactions on Medical
  Imaging}, vol.~13, no.~2, pp. 290--300, jun 1994. [Online]. Available:
  \url{http://dx.doi.org/10.1109/42.293921}
\BIBentrySTDinterwordspacing

\bibitem{Elbakri2002}
\BIBentryALTinterwordspacing
I.~Elbakri and J.~Fessler, ``Statistical image reconstruction for polyenergetic
  x-ray computed tomography,'' \emph{{IEEE} Transactions on Medical Imaging},
  vol.~21, no.~2, pp. 89--99, 2002. [Online]. Available:
  \url{http://dx.doi.org/10.1109/42.993128}
\BIBentrySTDinterwordspacing

\bibitem{Goldstein2009}
\BIBentryALTinterwordspacing
T.~Goldstein and S.~Osher, ``The split bregman method for l1-regularized
  problems,'' \emph{{SIAM} J. Imaging Sci.}, vol.~2, no.~2, pp. 323--343, jan
  2009. [Online]. Available: \url{http://dx.doi.org/10.1137/080725891}
\BIBentrySTDinterwordspacing

\bibitem{Boyd2010}
\BIBentryALTinterwordspacing
S.~Boyd, ``Distributed optimization and statistical learning via the
  alternating direction method of multipliers,'' \emph{{FNT} in Machine
  Learning}, vol.~3, no.~1, pp. 1--122, 2010. [Online]. Available:
  \url{http://dx.doi.org/10.1561/2200000016}
\BIBentrySTDinterwordspacing

\bibitem{Wang2012}
\BIBentryALTinterwordspacing
J.~Wang, J.~Ma, B.~Han, and Q.~Li, ``Split bregman iterative algorithm for
  sparse reconstruction of electrical impedance tomography,'' \emph{Signal
  Processing}, vol.~92, no.~12, pp. 2952--2961, dec 2012. [Online]. Available:
  \url{http://dx.doi.org/10.1016/j.sigpro.2012.05.027}
\BIBentrySTDinterwordspacing

\bibitem{Ramani2012}
\BIBentryALTinterwordspacing
S.~Ramani and J.~A. Fessler, ``A splitting-based iterative algorithm for
  accelerated statistical x-ray {CT} reconstruction,'' \emph{{IEEE}
  Transactions on Medical Imaging}, vol.~31, no.~3, pp. 677--688, mar 2012.
  [Online]. Available: \url{http://dx.doi.org/10.1109/TMI.2011.2175233}
\BIBentrySTDinterwordspacing

\bibitem{Chun2014}
\BIBentryALTinterwordspacing
S.~Y. Chun, Y.~K. Dewaraja, and J.~A. Fessler, ``Alternating direction method
  of multiplier for tomography with nonlocal regularizers,'' \emph{{IEEE}
  Transactions on Medical Imaging}, vol.~33, no.~10, pp. 1960--1968, oct 2014.
  [Online]. Available: \url{http://dx.doi.org/10.1109/TMI.2014.2328660}
\BIBentrySTDinterwordspacing

\bibitem{Defrise2006}
\BIBentryALTinterwordspacing
M.~Defrise, F.~Noo, R.~Clackdoyle, and H.~Kudo, ``Truncated hilbert transform
  and image reconstruction from limited tomographic data,'' \emph{Inverse
  Problems}, vol.~22, no.~3, p. 1037, 2006. [Online]. Available:
  \url{http://stacks.iop.org/0266-5611/22/i=3/a=019}
\BIBentrySTDinterwordspacing

\bibitem{Tikhonov1977}
\BIBentryALTinterwordspacing
A.-I.~N. Tikhonov, \emph{Solutions of Ill Posed Problems (Scripta series in
  mathematics)}.\hskip 1em plus 0.5em minus 0.4em\relax Vh Winston, 1977.
  [Online]. Available:
  \url{http://www.amazon.com/Solutions-Posed-Problems-Scripta-mathematics/dp/0470991240%3FSubscriptionId%3D0JYN1NVW651KCA56C102%26tag%3Dtechkie-20%26linkCode%3Dxm2%26camp%3D2025%26creative%3D165953%26creativeASIN%3D0470991240}
\BIBentrySTDinterwordspacing

\bibitem{Huber1964}
\BIBentryALTinterwordspacing
P.~J. Huber, ``Robust estimation of a location parameter,'' \emph{Ann. Math.
  Statist.}, vol.~35, no.~1, pp. 73--101, mar 1964. [Online]. Available:
  \url{http://dx.doi.org/10.1214/aoms/1177703732}
\BIBentrySTDinterwordspacing

\bibitem{Rudin1992}
\BIBentryALTinterwordspacing
L.~I. Rudin, S.~Osher, and E.~Fatemi, ``Nonlinear total variation based noise
  removal algorithms,'' \emph{Physica D: Nonlinear Phenomena}, vol.~60, no.
  1-4, pp. 259--268, nov 1992. [Online]. Available:
  \url{http://dx.doi.org/10.1016/0167-2789(92)90242-F}
\BIBentrySTDinterwordspacing

\bibitem{Peters1981}
\BIBentryALTinterwordspacing
T.~M. Peters, ``Algorithms for fast back- and re-projection in computed
  tomography,'' \emph{{IEEE} Trans. Nucl. Sci.}, vol.~28, no.~4, pp.
  3641--3647, 1981. [Online]. Available:
  \url{http://dx.doi.org/10.1109/TNS.1981.4331812}
\BIBentrySTDinterwordspacing

\bibitem{Zhuang1994}
\BIBentryALTinterwordspacing
W.~Zhuang, S.~Gopal, and T.~Hebert, ``Numerical evaluation of methods for
  computing tomographic projections,'' \emph{{IEEE} Trans. Nucl. Sci.},
  vol.~41, no.~4, pp. 1660--1665, 1994. [Online]. Available:
  \url{http://dx.doi.org/10.1109/23.322963}
\BIBentrySTDinterwordspacing

\bibitem{Zeng1993}
\BIBentryALTinterwordspacing
G.~Zeng and G.~Gullberg, ``A ray-driven backprojector for backprojection
  filtering and filtered backprojection algorithms,'' in \emph{1993 {IEEE}
  Conference Record Nuclear Science Symposium and Medical Imaging
  Conference}.\hskip 1em plus 0.5em minus 0.4em\relax Institute of Electrical
  {\&} Electronics Engineers ({IEEE}). [Online]. Available:
  \url{http://dx.doi.org/10.1109/NSSMIC.1993.701833}
\BIBentrySTDinterwordspacing

\bibitem{Basu2002}
\BIBentryALTinterwordspacing
B.~D. Man and S.~Basu, ``Distance-driven projection and backprojection,'' in
  \emph{2002 {IEEE} Nuclear Science Symposium Conference Record}.\hskip 1em
  plus 0.5em minus 0.4em\relax Institute of Electrical {\&} Electronics
  Engineers ({IEEE}). [Online]. Available:
  \url{http://dx.doi.org/10.1109/NSSMIC.2002.1239600}
\BIBentrySTDinterwordspacing

\bibitem{Basu2004}
\BIBentryALTinterwordspacing
------, ``Distance-driven projection and backprojection in three dimensions,''
  \emph{Physics in Medicine and Biology}, vol.~49, no.~11, p. 2463, 2004.
  [Online]. Available: \url{http://stacks.iop.org/0031-9155/49/i=11/a=024}
\BIBentrySTDinterwordspacing

\bibitem{Chapman1981}
\BIBentryALTinterwordspacing
C.~H. Chapman, ``Generalized radon transforms and slant stacks,''
  \emph{Geophysical Journal International}, vol.~66, no.~2, pp. 445--453, aug
  1981. [Online]. Available:
  \url{http://dx.doi.org/10.1111/j.1365-246X.1981.tb05966.x}
\BIBentrySTDinterwordspacing

\bibitem{Toft1996}
P.~Toft and J.~Sørensen, ``The radon transform - theory and implementation,''
  Ph.D. dissertation, 11 1996.

\bibitem{YongLong2010}
\BIBentryALTinterwordspacing
Y.~Long, J.~A. Fessler, and J.~M. Balter, ``3d forward and back-projection for
  x-ray {CT} using separable footprints,'' \emph{{IEEE} Transactions on Medical
  Imaging}, vol.~29, no.~11, pp. 1839--1850, nov 2010. [Online]. Available:
  \url{http://dx.doi.org/10.1109/TMI.2010.2050898}
\BIBentrySTDinterwordspacing

\bibitem{Palenstijn2011}
\BIBentryALTinterwordspacing
W.~Palenstijn, K.~Batenburg, and J.~Sijbers, ``Performance improvements for
  iterative electron tomography reconstruction using graphics processing units
  ({GPUs}),'' \emph{Journal of Structural Biology}, vol. 176, no.~2, pp.
  250--253, nov 2011. [Online]. Available:
  \url{http://dx.doi.org/10.1016/j.jsb.2011.07.017}
\BIBentrySTDinterwordspacing

\bibitem{Papenhausen2013}
\BIBentryALTinterwordspacing
E.~Papenhausen, Z.~Zheng, and K.~Mueller, ``Creating optimal code for
  {GPU}-accelerated {CT} reconstruction using ant colony optimization,''
  \emph{Med. Phys.}, vol.~40, no.~3, p. 031110, 2013. [Online]. Available:
  \url{http://dx.doi.org/10.1118/1.4773045}
\BIBentrySTDinterwordspacing

\bibitem{Andersson2016}
\BIBentryALTinterwordspacing
F.~Andersson, M.~Carlsson, and V.~V. Nikitin, ``Fast algorithms and efficient
  {GPU} implementations for the radon transform and the back-projection
  operator represented as convolution operators,'' \emph{{SIAM} J. Imaging
  Sci.}, vol.~9, no.~2, pp. 637--664, jan 2016. [Online]. Available:
  \url{http://dx.doi.org/10.1137/15M1023762}
\BIBentrySTDinterwordspacing

\bibitem{Basu2000}
\BIBentryALTinterwordspacing
S.~Basu and Y.~Bresler, ``O(n/sup 2/log/sub 2/n) filtered backprojection
  reconstruction algorithm for tomography,'' \emph{{IEEE} Transactions on Image
  Processing}, vol.~9, no.~10, pp. 1760--1773, 2000. [Online]. Available:
  \url{http://dx.doi.org/10.1109/83.869187}
\BIBentrySTDinterwordspacing

\bibitem{Matej2004}
\BIBentryALTinterwordspacing
S.~Matej, J.~Fessler, and I.~Kazantsev, ``Iterative tomographic image
  reconstruction using fourier-based forward and back-projectors,''
  \emph{{IEEE} Transactions on Medical Imaging}, vol.~23, no.~4, pp. 401--412,
  apr 2004. [Online]. Available:
  \url{http://dx.doi.org/10.1109/TMI.2004.824233}
\BIBentrySTDinterwordspacing

\bibitem{Arcadu2016}
\BIBentryALTinterwordspacing
F.~Arcadu, M.~Nilchian, A.~Studer, M.~Stampanoni, and F.~Marone, ``A forward
  regridding method with minimal oversampling for accurate and efficient
  iterative tomographic algorithms,'' \emph{{IEEE} Transactions on Image
  Processing}, vol.~25, no.~3, pp. 1207--1218, mar 2016. [Online]. Available:
  \url{http://dx.doi.org/10.1109/TIP.2016.2516945}
\BIBentrySTDinterwordspacing

\bibitem{Natterer2001}
\BIBentryALTinterwordspacing
F.~Natterer, \emph{The Mathematics of Computerized Tomography (Classics in
  Applied Mathematics)}.\hskip 1em plus 0.5em minus 0.4em\relax SIAM: Society
  for Industrial and Applied Mathematics, 2001. [Online]. Available:
  \url{http://www.amazon.com/Mathematics-Computerized-Tomography-Classics-Applied/dp/0898714931%3FSubscriptionId%3D0JYN1NVW651KCA56C102%26tag%3Dtechkie-20%26linkCode%3Dxm2%26camp%3D2025%26creative%3D165953%26creativeASIN%3D0898714931}
\BIBentrySTDinterwordspacing

\bibitem{Siddon1985}
\BIBentryALTinterwordspacing
R.~L. Siddon, ``Fast calculation of the exact radiological path for a
  three-dimensional {CT} array,'' \emph{Med. Phys.}, vol.~12, no.~2, p. 252,
  1985. [Online]. Available: \url{http://dx.doi.org/10.1118/1.595715}
\BIBentrySTDinterwordspacing

\bibitem{Kak2001}
\BIBentryALTinterwordspacing
A.~C. Kak and M.~Slaney, \emph{Principles of Computerized Tomographic
  Imaging}.\hskip 1em plus 0.5em minus 0.4em\relax Society for Industrial {\&}
  Applied Mathematics ({SIAM}), jan 2001. [Online]. Available:
  \url{http://dx.doi.org/10.1137/1.9780898719277}
\BIBentrySTDinterwordspacing

\bibitem{Lyra2011}
\BIBentryALTinterwordspacing
M.~Lyra and A.~Ploussi, ``Filtering in {SPECT} image reconstruction,''
  \emph{International Journal of Biomedical Imaging}, vol. 2011, pp. 1--14,
  2011. [Online]. Available: \url{http://dx.doi.org/10.1155/2011/693795}
\BIBentrySTDinterwordspacing

\bibitem{Shepp1974}
\BIBentryALTinterwordspacing
L.~A. Shepp and B.~F. Logan, ``The fourier reconstruction of a head section,''
  \emph{{IEEE} Trans. Nucl. Sci.}, vol.~21, no.~3, pp. 21--43, jun 1974.
  [Online]. Available: \url{http://dx.doi.org/10.1109/TNS.1974.6499235}
\BIBentrySTDinterwordspacing

\bibitem{HuynhThu2008}
\BIBentryALTinterwordspacing
Q.~Huynh-Thu and M.~Ghanbari, ``Scope of validity of {PSNR} in image/video
  quality assessment,'' \emph{Electron. Lett.}, vol.~44, no.~13, p. 800, 2008.
  [Online]. Available: \url{http://dx.doi.org/10.1049/el:20080522}
\BIBentrySTDinterwordspacing

\end{thebibliography}


\end{document}